\documentclass{article}

\usepackage{microtype}
\usepackage{graphicx}
\usepackage{subfigure}
\usepackage{booktabs} % for professional tables
\usepackage{breqn}
\usepackage{footnote}
\usepackage{graphicx}
\usepackage{tipa}
\usepackage{bm}
\usepackage{paralist}
\usepackage{calrsfs}
\usepackage{subfiles}
\usepackage{xcolor}
\usepackage{mwe}
\usepackage{dutchcal}
\usepackage{upgreek}
\usepackage{bbm}
\usepackage{hyperref}

\usepackage[accepted]{icml2025}

\usepackage{amsmath}
\usepackage{amssymb}
\usepackage{mathtools}
\usepackage{amsthm}

\usepackage[capitalize,noabbrev]{cleveref}

\theoremstyle{plain}
\newtheorem{theorem}{Theorem}[section]
\newtheorem{proposition}{Proposition}[section]
\newtheorem{lemma}{Lemma}[section]

\theoremstyle{definition}
\newtheorem{definition}[theorem]{Definition}
\newtheorem{assumption}[theorem]{Assumption}
\theoremstyle{remark}
\newtheorem{remark}{Remark}

\usepackage[textsize=tiny]{todonotes}

\newcommand{\R}{\mathbb{R}}

\newcommand{\bb}[1]{\mathbb{#1}}

\newcommand{\wh}[1]{\hat{#1}}

\newcommand{\bfa}[1]{\boldsymbol{#1}}

\DeclareMathAlphabet{\pazocal}{OMS}{zplm}{m}{n}
\newcommand{\ca}[1]{\pazocal{#1}}

\newcommand{\lef}{\left}
\newcommand{\rig}{\right}

\newcommand{\tda}[1]{\tilde{\bfa{#1}}}

\newcommand{\sign}{\text{sign}}
\newcommand{\pta}{\partial}

\newcommand{\mbbm}[1]{\mathbbm{#1}}

\newcommand{\diag}{\text{diag}}

\newcommand{\ub}[1]{\underbrace{#1}}

\newcommand{\bl}{{\bigg (}}
\newcommand{\br}{{\bigg )}}

\newcommand{\sigmoid}{\text{Sigmoid}}

\newcommand{\sm}[1]{\small{#1}\normalsize}
\newcommand{\tny}[1]{\footnotesize{#1}\normalsize}
\newcommand{\ttny}[1]
{\scriptsize{#1}\normalsize}
\newcommand{\softmaxx}{\text{Softmax}}
\newcommand{\softmax}{\text{SoftMax}}

\DeclareMathOperator*{\argmin}{arg\,min}
\newcommand{\wha}[1]{\wh{\bfa #1}}
\newcommand{\vertiii}[1]{{\left\vert\kern-0.25ex\left\vert\kern-0.25ex\left\vert #1 \right\vert\kern-0.25ex\right\vert\kern-0.25ex\right\vert}}
\newcommand{\tfp}{\text{Transformer+}}

\icmltitlerunning{Transformers and EM Algorithm}

\begin{document}

\twocolumn[
\icmltitle{Transformers versus the EM Algorithm in Multi-class Clustering}



% \icmlsetsymbol{}{}

\begin{icmlauthorlist}
\icmlauthor{Yihan He}{pu}
\icmlauthor{Hong-Yu Chen}{nwu}
\icmlauthor{Yuan Cao}{hku}
\icmlauthor{Jianqing Fan}{pu}
\icmlauthor{Han Liu}{nwu}

\end{icmlauthorlist}

\icmlaffiliation{pu}{Princeton University, Princeton, NJ, USA}
\icmlaffiliation{nwu}{Northwestern University, Evanston, IL, USA}
\icmlaffiliation{hku}{The University of Hong Kong, Hong Kong}

\icmlcorrespondingauthor{Yihan He}{yihan.he@princeton.edu}

\icmlkeywords{Machine Learning}
\vskip 0.3in
]



\printAffiliationsAndNotice{} % otherwise use the standard text.

\begin{abstract}
LLMs demonstrate significant inference capacities in complicated machine learning tasks, using the Transformer model as its backbone. Motivated by the limited understanding of such models on the unsupervised learning problems, we study the learning guarantees of Transformers in performing multi-class clustering of the Gaussian Mixture Models. We develop a theory drawing strong connections between the Softmax Attention layers and the workflow of the EM algorithm on clustering the mixture of Gaussians. Our theory provides approximation bounds for the Expectation and Maximization steps by proving the universal approximation abilities of multivariate mappings by Softmax functions. In addition to the approximation guarantees, we also show that with a sufficient number of pre-training samples and an initialization, Transformers can achieve the minimax optimal rate for the problem considered. Our extensive simulations empirically verified our theory by revealing the strong learning capacities of Transformers even beyond the assumptions in the theory, shedding light on the powerful inference capacities of LLMs.
\end{abstract}

\section{Introduction}
Large Language Models (LLMs) have demonstrated significant success in learning and performing inference on real world high dimensional datasets. Most modern LLMs use the Transformer model \citep{vaswani2017attention} as their backbone. 

Many existing works have considered the theoretical guarantees of the in-context-learning setup of Transformers \citep{bai2024transformers, akyurek2022learning}. However, in practice, LLMs require a significant amount of pretraining data to achieve their empirical advantage. And, little is known about the unsupervised learning guarantees of Transformers, especially after a sufficient number of problem instances are observed by the Transformer model in the pre-training phase. Motivated by the strong empirical performance of Transformers, we provide theoretical analysis of the Transformers on a standard unsupervised learning problem of clustering a mixture of Gaussians in the multi-class setup. Our results suggest that Transformers, like human brains, can benefit from experienced problem instances and learn the way to solve the problem (algorithms). Then, when fed with a new problem instance, Transformers can solve it through the learned algorithms naturally.

The problem of clustering a mixture of multivariate Gaussian is one of the most standard unsupervised learning problems \citep{bishop2006pattern} that can be solved by the EM algorithm or Lloyd's algorithm \citep{lloyd1982least}. The EM algorithm contains both the \emph{Expectation} and the \emph{Maximization} sub-procedures where the \emph{Expectation Step} creates a function for \textbf{the expectation} of the log-likelihood evaluated using the current estimate for the parameters and the \emph{Maximization Step} computes parameters maximizing the expected log-likelihood given by the \emph{Expectation Step}. We draw connections between the Softmax Attention in Transformers and the EM algorithms through the following:
\begin{enumerate}
    \item \emph{The Expectation Step} involves a normalized sum in \textbf{the expectation} whose weight vector is naturally given by the output of the softmax function as 
    $\begin{bmatrix}
        \frac{\exp(z_1)}{\sum_{i=1}^D\exp(z_i)},\ldots,\frac{\exp(z_D)}{\sum_{i=1}^D\exp(z_i)}
    \end{bmatrix}$.
    \item \emph{The Maximization Step} involves finding the index with the maximum value in a vector (the Hardmax Function). This is naturally approximated by the Softmax function as its name suggests.
\end{enumerate}
Given the strong connections of the two steps to the Softmax function, we build an approximation theory for Lloyd's algorithm in a constructive manner. We also note that existing works only build approximation bound for multihead ReLU neural networks \citep{bach2017breaking} while the Softmax approximation of multivariate to multivariate mapping remains a myth. We resolve this obstacle by proving an approximation bound for multi-head Transformers on a class of $\bb R^{d_1}\to\bb R^{d_2}$ mappings that might be of independent interests.

\paragraph{Contributions.} We summarize our major contributions as follows:
\begin{enumerate}
    \item We rigorously show that a pre-trained Transformer can perform multi-class clustering by drawing its connection to Lloyd's algorithm, which is used as a proof machine. We provide constructive proof and error bound for the approximation;
    \item We further consider the setup where the Transformer model is trained with independent instances from a class of clustering problems whose labels are used as supervision. We show that Transformers are able to generalize the mapping on new clustering problem instances. We provide upper bounds on the generalization error for the empirical risk minimizer in the pre-training task. Moreover, we show that given a sufficient number of training instances and proper initialization, pre-trained Transformers reach the fundamental limit of the problem;
    \item We systematically evaluate the performance of Transformers through extensive simulations. These empirical results demonstrate that Transformers perform well in the multi-class clustering task even when the assumptions leading to the theoretical results no longer hold.
\end{enumerate}

\subsection{Related Works}
\paragraph{Transformers are algorithm approximators.} Recently, the capacity of Transformers to automatically performing certain algorithms has drawn great attention from researchers. In particular, a rich line of recent works studied the expressive power of Transformers to perform in-context learning (ICL) \citep{akyurek2022learning,bai2024transformers,abernethy2024mechanism,li2023transformers2,jeon2024informationtheoretic}. Specifically, \citet{akyurek2022learning,bai2024transformers,abernethy2024mechanism} studied how Transformers perform gradient descent based training to perform ordinary or sparse linear regression on the context. \citet{chen2024transformers} further showed how Transformers utilize the multi-head structure to perform in-context sparse linear regression.  \citet{li2023transformers2} studied the generalization and stability of transformers in ICL tasks. \citet{jeon2024informationtheoretic} studies the information-theoretical lower bound of in-context learning. Another closely related line of works studied how Transformers can be pretrained by gradient descent to perform certain tasks. Specifically, \citet{zhang2023trained,huang2023context,chen2024training} studied the pretraining optimization dynamics of Transformers to learn in-context linear prediction rules. \citet{li2024one} showed that one-layer Transformers can be trained to perform one-nearest neighbor classification in context. \citet{ahn2024transformers,giannou2024well} studied the training of Transformers in learning various optimization methods. \citet{li2023transformers}  studied how Transformers can be trained to learn topic models. \citet{jelassi2022vision} proved that Vision Transformers can learn a class of image-like data whose patches follow certain spatial structures. \citet{zhang2025transformer} studied how Transformers can learn to perform variable selection in ``group-sparse'' linear regression.
% \paragraph{In-Context Learning with Transformers.} Recent studies have explored the in-context learning (ICL) capabilities of Transformers \citep{garg2022can, bai2024transformers}. Specifically, \citet{bai2024transformers} examined the approximation and generalization properties of Transformers in ICL tasks, including various linear and logistic regression scenarios. \citet{akyurek2022learning, von2023transformers} investigated how Transformers approximate gradient descent during ICL. In our work, we leverage the Power Method to provide theoretical guarantees for the approximation capabilities of Transformers. Notably, obtaining eigenvectors through gradient descent is challenging due to the lack of an explicit functional form. To our knowledge, this is the first study to offer theoretical guarantees for Transformers' approximation of the Power Method. For further practical insights into ICL, see \citet{dong2022survey} and the references therein.

\paragraph{Other theoretical studies on Transformers.} Various efforts have been made to gain a theoretical understanding of Transformers. \citet{yun2019transformers} analyzed the universal approximation properties of Transformers for sequence-to-sequence functions. \citet{li2023transformers} studied the mean-filed limit of large-scale Transformers and proved global convergence in regression tasks. 
\citet{perez2021attention} showed that Transformers with hard-attention are Turing complete exclusively based on their capacity to compute and access internal dense representations of the data. \citet{bhattamishra2020computational}
 further provided an alternate and simpler proof to show that vanilla Transformers are Turing-complete, and then proved that Transformers with only positional masking and without any positional encoding are also Turing-complete. \citet{liu2022transformers} showed that a low-depth Transformer can represent the computations of any finite-state automaton by hierarchically reparameterizing its recurrent dynamics. \citet{yao2021self} demonstrated that Transformers can efficiently process bounded hierarchical languages, offering better space complexity compared to recurrent neural networks.

\paragraph{Notations}
In this work we follow the following notation conventions. The vector-valued variable is given by boldfaced characters. We denote $[n]:=\{1,\ldots,n\}$ and $[i:j]:=\{i,i+1,\ldots, j\}$ for $i<j$. The universal constants are given by $C$ and are ad hoc. For a vector $\bfa v$ we denote $\Vert\bfa v\Vert_2$ as its $L_2$ norm. For a matrix $\bfa A\in\bb R^{m\times n}$ we denote its operator norm as $\Vert\bfa A\Vert_2:=\sup_{\bfa v\in\bb S^{n-1}}\Vert\bfa A\bfa v\Vert_2$. Given two sequences $a_n$ and $b_n$, we denote $a_n\lesssim b_n$ or $a_n = O(b_n)$ if $\limsup_{n\to\infty}|\frac{a_n}{b_n}|<\infty$ and $a_n=o(b_n)$ if $\limsup_{n\to\infty}|\frac{a_n}{b_n}|=0$. We denote $\mbbm 1_A$ as the indicator function for event $A$. The universal constants are denoted by $C$ in this work and are ad hoc. We use $\ca B(\Vert\cdot\Vert, r)$ to denote a ball with radius $r$ under the norm $\Vert\cdot\Vert$.  

\paragraph{Organizations} The rest of the paper is organized as follows: Section \ref{sect2} reviews standard contexts and describes the learning problem; Section \ref{sect3} provides rigorous theoretical results and sketches of proof; Section \ref{sect4} provides extensive experimental details and results; Section \ref{sect5} discusses the limitations and potential future works. The detailed proofs and additional figures in experiments are delayed to the appendix. The supplementary materials include the code for the experiments.

\section{Connecting Transformers with EM}\label{sect2}
This section discusses the connections between the EM algorithm and the Transformer architecture. 
Our discussion is split into $2$ separate subsections: In \ref{Transformers}, we review the mathematical definitions of the Softmax-based Transformer model; In \ref{emintro}, we review the EM algorithm and connect it with the multiphase Transformer design. In section \ref{pretrain}, we discuss the pretraining procedure of the Transformers.
\subsection{The Transformer Architecture}\label{Transformers}
We consider the Softmax Attention Layer, which is defined as follows:
\begin{definition}[Softmax Attention]\label{def:attention}
  The Softmax Attention layer is defined as a self-attention layer with $M$ heads denoted as $Attn_{\bfa\theta_1}(\cdot)$ with parameters $\bfa\theta_1=\{(\bfa V_m,\bfa Q_m,\bfa K_m)\}_{m\in[M]}\subset\bb R^{D\times D}$. On input sequence $\bfa H\in\bb R^{D\times N}$,
    {\begin{align*}
       Attn_{\bfa\theta_1}&(\bfa H)=
            \bfa H\\
            &+\sum_{m=1}^M(\bfa V_m\bfa H)\softmax\Big((\bfa Q_m\bfa H)^\top(\bfa K_m\bfa H)\Big),
    \end{align*}}
    where $\softmax$ is the activation function defined by
    \begin{align*}
        \softmax(\bfa x) = \begin{bmatrix}
            \frac{\exp(x_1)}{\sum_{i=1}^d\exp(x_i)}&\ldots&\frac{\exp(x_d)}{\sum_{i=1}^d\exp(x_d)}
        \end{bmatrix}^\top,
    \end{align*}
    for all $\bfa x\in\bb R^d$.
\end{definition}
In addition to the Softmax Attention layer, we also consider an un-normalized Attention layer, given by
\begin{definition}[Un-normalized Attention]\label{def:mattention}
  The un-normalized Attention layer is defined as a self-attention layer with $M$ heads and denoted as $nAttn_{\bfa\theta_1}(\cdot)$ with parameters $\bfa\theta_1=\{(\bfa V_m,\bfa Q_m,\bfa K_m)\}_{m\in[M]}\subset\bb R^{D\times D}$. On input sequence $\bfa H\in\bb R^{D\times N}$,
    {\begin{align*}
       nAttn_{\bfa\theta_1}&(\bfa H)=
            \bfa H+\sum_{m=1}^M(\bfa V_m\bfa H)(\bfa Q_m\bfa H)^\top(\bfa K_m\bfa H).
    \end{align*}}
\end{definition}
\begin{remark}
    The un-normalized Attention layer is the Attention layer without the non-linear activation function. This layer is studied mainly for technical reasons. We also provide results not using the un-normalized Attention layer, despite having weaker rates. 
\end{remark}
% \begin{remark}
%     We note that in the literature \citep{bai2024transformers} only considers the approximation guarantees yield by the ReLU Transformers. However, existing LLMs predominantely utilize the Softmax-based Attention layers. Our approximation guarantee for the Softmax Attention layers is new in the literature, relying on our new approximation result for the Softmax function, discussed in section \ref{sect31}. 

%     We also replace concatenation of multihead attentions by their average. Despite adaptations are employed for the technical convinience, in the simulation section we carefully evaluate the effect of these modifications on the performance of the model.
% \end{remark}
The following defines the classical Fully-Connected (FC) layers with residual connections.
\begin{definition}[FC Layer] A FC layer with hidden dimension $D^\prime$ is denoted as $FC_{\bfa\theta}(\cdot)$ with parameter $\bfa\theta_2\in(\bfa W_1,\bfa W_2)\in\bb R^{D^\prime\times D}\times\bb R^{D\times D^\prime}$. On any input sequence $\bfa H\in\bb R^{D\times N}$, we define
$$   FC_{\bfa\theta_2}(\bfa H):=\bfa H+\bfa W_2\sigma(\bfa W_1\bfa H).$$
\end{definition}
Then, we use the above definitions on the FC and the Attn/nAttn layers to define the Transformer model and the Transformer+ model.
\begin{definition}[Transformer]\label{transform}
    We define the Transformer $TF_{\bfa\theta}(\cdot)$ as a composition of the self-attention layers with the FC layers. Consider the output dimension to be $\tilde D$, a $L$-layered Transformer is defined by  \sm{\begin{align*}
        TF_{\bfa\theta}&(\bfa H) :=\\
        &\tda W_0\times FC_{\bfa\theta_{2}^L}(Attn_{\bfa\theta_{1}^L}(\cdots FC_{\bfa\theta_{2}^1}(Attn_{\bfa\theta_{1}^1}(\bfa H)))\times\tda W_1,
    \end{align*}}
    where $\tda W_0\in\bb R^{d_1\times D}$ and $\tda W_1\in\bb R^{N\times d_2}$.
\end{definition}
The two additional matrices $\tda W_0$ and $\tda W_1$ serve for the dimension adjustment purpose such that the output of $TF_{\bfa\theta}(\bfa H)$ or $TF^+_{\bfa\theta}(\bfa H)$ will be of dimension $\bb R^{d_1\times d_2}$.

Then, we introduce a class of models called the Transformer+, which includes the un-normalized Attention layer.
\begin{definition}[Transformer+]
    Under the same notations as definition \ref{transform}. We define the Transformer+ model $TF_{\bfa\theta}^+(\cdot)$ as
    \sm{\begin{align*}
        TF_{\bfa\theta}^+&(\bfa H) :=\tda W_0\times FC_{\bfa\theta_{2}^L}(A_{\bfa\theta_{1}^L}(\cdots FC_{\bfa\theta_{2}^1}(A_{\bfa\theta_{1}^1}(\bfa H)))\times\tda W_1,
    \end{align*}}
    where $A\in\{Attn,nAttn\}$ is either the Attn layer defined in definition 2.1 or the nAttn layer defined in definition 2.2.
\end{definition}

We use $\bfa \theta$ to denote all the parameters in the Transformer and the super-index $\ell$ to denote the parameter matrix corresponding to the $\ell$-th layer.
    Under such definition, the parameter $\bfa\theta$ is given by 
    \sm{\begin{align*}
        \bfa \theta = \{\{(\{\bfa Q_m^{(\ell)},\bfa K_m^{(\ell)},\bfa V_m^{(\ell)}\}_{m\in[M]}, \bfa W_{1}^{(\ell)},\bfa W_{2}^{(\ell)})\}_{\ell\in[L]},\tda W_{0},\tda W_{1}\}.
    \end{align*}}
Following the notations in \citep{bai2024transformers}, we define the operator norm of the parameter $\bfa\theta$ as follows.
    \begin{align*}
&\vertiii{\bfa\theta}:=\max_{\ell\in[L]}\Big\{\max_{m\in[M^{(\ell)}]}\lef\{\Vert \bfa Q_m^{(\ell)}\Vert_2,\Vert\bfa K_m^{(\ell)}\Vert_2\rig\}\\
&+\Vert\tda W_0\Vert_2+\Vert\tda W_1\Vert_2+\sum_{m=1}^{M^{(\ell)}}\Vert\bfa V_m^{(\ell)}\Vert_2+\Vert\bfa W_1^{(\ell)}\Vert_2+\Vert\bfa W_2^{(\ell)}\Vert_2\Big\}, 
    \end{align*}
    where $M^{(\ell)}$ is the number of heads of the $\ell$-th attention layer.
    It is also shown in \citep{bai2024transformers} that such a norm relates to the Lipschitz constant of Transformers, which controls the model complexity and leads to the generalization bound. Hence, in this work, we consider the following space of the model
    \begin{align*}
        \Theta(B_{\bfa\theta},B_{M},B_L) &= \bigg\{(\bfa\theta, \{M^{(\ell)}\}_{\ell\in[L]}, L): \vertiii{\bfa\theta}\leq B_{\bfa\theta},\\
        &\quad\quad  \sup_{\ell\in[L]}M^{(\ell)}\leq B_{M}, L\leq B_L \bigg\}.
    \end{align*}
    And for the subspace of $\bfa\theta$ given $M$ and $L$ as hyperparamaters, we denote by
    $\Theta_{B_M,B_L}(B_{\bfa\theta})$.

\subsection{The Learning Problem and EM}\label{emintro}
In this section, we first provide notations for the sub-Gaussian mixture models and the clustering problem. Then, we provide the literature on the EM Algorithm and Lloyd's algorithm.
\subsubsection{Clustering Mixture of Gaussians}
 We take samples $\{\bfa X_i\}_{i\in[N]}$ from a sub-Gaussian mixture model with in total of $k$ centers $\{\bfa\mu_i\}_{i\in[k]}$. In particular, we let
 \begin{align*}
  \bfa X:=\begin{bmatrix}
      \bfa X_1\ldots\bfa X_N
  \end{bmatrix},\quad   \bfa X_i := \bfa\mu_{z_i}+\bfa\omega_i\text{ for all }i\in[N],
 \end{align*}
 where $z_i:\in[k]$ corresponds to the membership of $i$-th index. We assume the following condition to hold for $\bfa\omega_i$.
 \begin{assumption}
     $\{\bfa\omega_i\}_{i\in[N]}$ are i.i.d. zero mean random variables from sub-Gaussian distribution that satisfies 
     $\bb E[\exp(\bfa a^\top\bfa\omega)]\leq\exp\lef(\frac{1}{2}\sigma^2\Vert\bfa a\Vert_2^2\rig)$ for all $\bfa a\in\bb R^d$.
 \end{assumption}
 We consider the mapping from $z$ to a set of one-hot vectors
\sm{\begin{align}\label{defp}
    \bfa P_1(z):=\begin{bmatrix}
        \bfa p_{1,1}&\bfa p_{1,2}&\ldots&\bfa p_{1,N}
    \end{bmatrix}\in\bb R^{k\times N},\bfa p_{1,i,j}:=\mbbm 1_{j=z_i}.
\end{align}}

Define $\ca S_k:[k]\to[k]$ as the set of permutations of $[k]$. We consider the following loss function for the Transformer output. 
\begin{align*}
    L(A_{\bfa\theta}(\bfa H),\bfa P_1(\bfa z)):=\inf_{\pi\in\ca S_k}\frac{1}{N}\Vert\bfa P_1(\pi(\bfa z_i))-A_{\bfa\theta}(\bfa H)\Vert_{1,1},
\end{align*}
where $A\in\{TF,TF^+\}$. 
\paragraph{The Parameter Space} This work considers the following space of parameters of the generative model. Here, we denote $F_{\omega}$ as the distribution of the random variable $\omega$.
\begin{align*}
    \Theta_{GM}&=\Big\{(\bfa\mu,\bfa z,F_{\bfa \omega}),\bfa\mu\in\bb R^{d\times k}, \Delta\leq \min_{i\neq j}\Vert\bfa\mu_i-\bfa\mu_j\Vert_2,\\
    &\bfa z:[N]\to[k], |\{i\in[N],\bfa z_i=u\}|\geq\alpha n, \forall u\in[k], \\
    &\omega_i\text{ is i.i.d. }\sigma\text{ sub-Gaussian random variable } \forall i\in[N]\Big\}.
\end{align*}
We further consider the solution space $\Theta_{\bfa A}=\{\bfa A:\sum_{j=1}^N\bfa A_{ij}=1,\forall i\in[k],\bfa A\in[0,1]^{k\times N}\}$.
Then, the fundamental limit of the problem class $\Theta_{GM}$ is given by the following lemma.
\begin{lemma}[Lower Bound \citep{yu2015useful}]\label{minimaxlb} For model class $\Theta_{GM}$, given $\frac{\Delta}{\sigma \log(k/\alpha)}\to \infty$, 
\begin{align*}
    \inf_{\wha A\in\Theta_{\bfa A}}\sup_{(z,\theta,F_{\omega})}\bb E[L(\wha A,\bfa P_1)]\geq\exp\lef(-(1+o(1))\frac{\Delta^2}{8\sigma^2} \rig).
\end{align*}
\end{lemma}
\begin{remark}
    The above result implies that the difficulty of this problem is governed by the Signal-to-Noise ratio $\frac{\Delta}{\sigma}$. In particular, the above results imply that the minimax rate of this problem is largely dependent on the distance between the two closest centroids.
    We also note that the original result is instead on the $0-1$ loss between $\wha z$ and $\bfa z$. However, it is also not difficult to show the same results hold for the solution space $\Theta_{\bfa A}$ and our defined loss $L$.
\end{remark}

\subsubsection{The EM (Lloyd's) Algorithm}
Lloyd's algorithm is a special case of EM algorithm on the Gaussian mixture model, which is formally stated by Algorithm \ref{alg:loyld}. The Lloyd's algorithm iteratively updates: \textbf{(1)} The centroid of each cluster; \textbf{(2)} The membership of each sample. Since Lloyd's algorithm requires an initial input $\{\wha\mu_i^{(0)}\}$, \citet{lu2016statistical} has shown that given a proper initialization algorithm \ref{alg:example}, Lloyd's algorithm provably achieves good performance. An example initialization algorithm is given by algorithm \ref{alg:example} where the spectral algorithm and k-means++ algorithm \citep{kumar2004simple} are first called to obtain approximate solutions.
\begin{algorithm}[h]
   \caption{Lloyd's Algorithm}
   \label{alg:loyld}
\begin{algorithmic}
   \STATE {\bfseries Input:} A sample matrix from Mixture of Gaussians $\bfa X\in\bb R^{d\times N}$, number of iterations $\tau$, and initial centroids $\{\wha\mu^{(0)}_i\}_{i\in[k]}$.
   \STATE Compute the Initial Clusters 
   \begin{align}\label{z0}
       \wha z^{(0)}_i = \argmin_{i\in[k]}\Vert \bfa X_j - \wha\mu_i^{(0)}\Vert_2\quad\text{for all }j\in[N].
   \end{align}
   \FOR{$\ell=1$ {\bfseries to} $\tau$}
   \STATE \textbf{(1) The Expectation Step:} Update the centroid by 
   \sm{\begin{align*}
       \wha \mu_i^{(\ell)}=\frac{\sum_{j=1}^N\mbbm 1_{\wha z_j^{(\ell-1)}=i}\bfa X_j}{\sum_{j=1}^N\mbbm 1_{\wha z_j^{(\ell-1)}=i}}.
   \end{align*}}
   \STATE \textbf{(2) The Maximization Step:} Update the cluster assignment by
   \begin{align*}
       \wha z^{(\ell)}_j=\argmin_{i\in[k]}\Vert \bfa X_j-\wha\mu_i^{(\ell)}\Vert_2\quad\text{for all }j\in[N].
   \end{align*}
   \ENDFOR
\end{algorithmic}
\end{algorithm}

\subsection{Pretraining with Supervised Learning}\label{pretrain}
The clustering problem is unsupervised where no labels are given. Transformers are usually used in the supervised learning setup. To let Transformers learn the algorithms, we perform supervised pre-training. 

In this setup, we are first given in a total of $n$ pretraining instances $\{\bfa X^{(i)}\}_{i\in[n]}$ and $\{\bfa z^{(i)}\}_{i\in[n]}$. We also form the pretraining instances by feeding the Transformer with the initialization given by \ref{alg:example}, encoded in $\{\bfa H^{(i)}\}_{i\in[n]}$. Then, we train the Transformer using the standard supervised learning on this set. Since the optimization of Transformers is non-convex and difficult to analyze, we consider the empirical risk minimizer of the Transformer given by 
\begin{align}\label{ERM}
    \wha\theta :=\argmin_{\bfa\theta\in\Theta_{(B_M,B_L)}(B_{\bfa\theta})}\sum_{i=1}^nL\lef(A_{\bfa\theta}(\bfa H^{(i)}),\bfa P_1(\bfa z^{(i)})\rig),
\end{align}
where $A\in\{TF,TF^+\}$.

In our theoretical analysis, we construct the input of the Transformer as a \emph{context-augmented matrix} given by the following
\sm{\begin{align}\label{aux}
    \bfa H=\begin{bmatrix}
        \bfa X\\
        \bfa P
    \end{bmatrix}, \bfa P=\begin{bmatrix}
        \wha \mu_{1}^{(0)}&\wha \mu_{2}^{(0)}&\ldots&\wha \mu_{k}^{(0)}&\ldots&\bfa 0\\
        \bfa p_{1,1}^{(0)}&\bfa p_{1,2}^{(0)}&\ldots&\bfa p_{1,k}^{(0)}&\ldots&\bfa p_{1,N}^{(0)}\\
        \bfa p_{2,1}&\bfa p_{2,2}&\ldots&\bfa p_{2,k}&\ldots&\bfa p_{2,N}\\
        1&1&\ldots&1&\ldots&1\\
        &&\bfa 0&&&
    \end{bmatrix},
\end{align}}
where $\bfa H\in\bb R^{D\times N}$ and $\bfa P^{(D-d)\times N}$. We let the input dimension $D\leq Ckd$ for some universal constant $C$.
The matrix $\bfa P$ contains contextual information. For the first row, $\{\wha\mu_i^{(0)}\}_{i\in[k]}\subset\bb R^{d}$ are the initial centroid estimates given by the initialization algorithm \ref{alg:example}. Then the next row $\{\bfa p_{1,i}\}_{i\in[N]}\subset[0,1]^{k}$ is given by $\bfa P_1(\wha z^{(0)})$ as in \eqref{defp} where $\wha z^{(0)}$ corresponds to the initialized membership in \eqref{z0}. Then the row $\{\bfa p_{2,i}\}_{i\in[N]}\subset\bb R^d$ satisfies
\begin{align*}
    \bfa p_{2,i,j}=\mbbm 1_{i=j}\text{ if }j\in[d].
\end{align*}
And the last row is set to all $1$ for the technical purpose of introducing constants into the Softmax function. 
\begin{algorithm}[tb]
   \caption{Initialization by Spectral Clustering}
   \label{alg:example}
\begin{algorithmic}
   \STATE {\bfseries Input:} Matrix $\bfa X\in\bb R^{d\times N}$.
   \STATE Perform PCA on $\bfa X\bfa X^\top$ and obtain its top-k eigenvectors $\{\bfa V_i\}_{i\in[k]}$.
   \STATE Project the input matrix by $\tda X=\bfa V^\top \bfa X$.
   \STATE Solve the $k$-means program given by 
   \begin{align*}
       \tda z:=\argmin_{\wh z:[N]\to [k]}\min_{\{\bfa\mu_i\}_{i\in[N]}}\sum_{i=1}^k\Vert\bfa \mu_{\wh z_i}- \tda X_i\Vert_2
   \end{align*}
   by the $k$-means++ algorithm \citep{kumar2004simple}.
   \STATE{\bfseries Return:} Initial Cluster Assignment $\tda z$.
\end{algorithmic}
\end{algorithm}

% \paragraph{The Learning Problem.} Consider a set of samples $\{\bfa X_i\}_{i\in[u]}$ i.i.d. sampled from some distribution $p_{\bfa X}$, we construct their oracle top-$k$ principle components as $\bfa V_i=\begin{bmatrix}
%     \bfa v^{i,\top}_1&\ldots&\bfa v_k^{i,\top}
% \end{bmatrix}^\top$
% and the context-augmented input matrix as $\bfa H_i$ for each $\bfa X_i$. Then, the pretraining procedure is given by minimizing the following objective for some convex loss function $L(\cdot,\cdot):\bb R^{dk}\times \bb R^{dk}\to\bb R$,
% \begin{align}\label{ERM}
%  \wha\theta =\argmin_{\bfa\theta\in\Theta(B_{\bfa\theta}, B_M)}\sum_{i=1}^uL(TF_{\bfa\theta}(\bfa H_i), \bfa V_i).
% \end{align}
% Here we consider $\Theta(B_{\bfa\theta}):=\{\bfa\theta:\Vert\bfa\theta\Vert\leq B_{\bfa\theta}, \max_{\ell}M^{(\ell)}\leq B_M\}$ to be the space of parameters. We also consider guarantees in the $L_2$ norm which states that $L(\bfa x_1,\bfa x_2):=\Vert\bfa x_1-\bfa x_2\Vert_2$ in the theoretical part.
% Since $\wha\theta$ given by minimizing the empirical risk is not obtainable in practice, our theory only gives guarantee on the empirical risk minimizer. We further show that the local minimizers obtained through stochastic gradient optimization achieve good empirical performance in section \ref{sect4}.

\section{Theoretical Results}\label{sect3}
This section presents our theoretical results and a proof sketch of our results. This section is divided into three parts: section \ref{approx1} provides the approximation bound to EM algorithms by Transformers; section \ref{approx2} provides the generalization bound; section \ref{approx3} provides a short proof sketch over the main theorem.

\subsection{The Approximation Bound}\label{approx1}
We first present an approximation bound for Lloyd's algorithm by both the Transformer and the Transformer+.
\begin{theorem}\label{thm_main1}
   Assume that $d=o(N)$, $k<d$. There exists a Transformer with number of layers $L=\tau(3+3k)$ and norm $\vertiii{\bfa\theta}\lesssim C^d(\log N +M)$ with the number of heads $M$ such that
    \begin{align*}
        \Big\Vert TF_{\bfa\theta}&(\bfa H)-\begin{bmatrix}
            \bfa p_{1,1}^{(\tau)}&\bfa p_{1,2}^{(\tau)}&\ldots&\bfa p_{1,N}^{(\tau)}
        \end{bmatrix}\Big\Vert_2\\
        &\lesssim \tau C^d\Big(\sqrt k\sup_{j\in[k],\ell\in[\tau]}\Vert\wha\mu_j^{(\tau)}\Vert_2\sqrt{\frac{\log M}{M}}+N^{-1}\Big)
    \end{align*}
where $\tau$ is the number of iterations in the Lloyd's algorithm, and $p_{1,i}^{(\tau)}$ the one-hot coding of the membership $\wha z_i^{(\tau)}$ there.
\end{theorem}

\begin{remark}
The two terms in the bound come from the expectation and the maximization steps, respectively. The $N^{-1}$ term comes from the expectation step, where we relate the weighted average with the weights given by the Softmax function. The second term related to multi-head attention comes from a few multi-head approximation layers for some general functions. To achieve this bound, we prove a new result on the universal approximation of the Softmax function, which brings in the approximation term discussed in section \ref{approx3}. We also note that the $N^{-1}$ term can be further improved by introducing the non-activated Attention layer, given by the next theorem for Transformer+ architecture. 
\end{remark} 
\begin{theorem}\label{thm_main2}
    Under the same condition as theorem \ref{thm_main1}, there exists a \tfp with the number of layers $L=\tau(7+3k)$ and norm $\vertiii{\bfa\theta}\lesssim C^dM\log N$ with the number of heads $M$ such that
    \sm{\begin{align*}
        &\Big\Vert TF^+_{\bfa\theta}(\bfa H)-\begin{bmatrix}
            \bfa p_{1,1}^{(\tau)}&\bfa p_{1,2}^{(\tau)}&\ldots&\bfa p_{1,N}^{(\tau)}
        \end{bmatrix}\Big\Vert_2\lesssim \tau \Big(dN^{-100}\\
        &+C^d\sqrt k\sup_{j\in[k],\ell\in[\tau]}\Vert\wha\mu_j^{(\ell)}\Vert_2\sqrt{\frac{\log M}{M}}+C^d\sqrt{\frac{d^2\log M}{M}}\Big).
    \end{align*}}
\end{theorem}
\begin{remark}
    The key difference between Theorems \ref{thm_main1} and \ref{thm_main2} comes from different designs of the Expectation Step. Through the introduction of the non-activated Attention layer, we manage to approximate a wider range of functions, including the polynomials on $\bfa H$ with degree $3$. We also believe that the non-activated Attention layer is unnecessary if a stronger \emph{right-product} universal approximation result can be proved for the Softmax functions, which is discussed in section \ref{approx3}.
\end{remark}
\subsection{The Generalization Bounds}\label{approx2}
Given the approximation error provided by Theorems \ref{thm_main1} and \ref{thm_main2}, we further provide the generalization error bound for the ERM defined by \eqref{ERM}. We consider the problem instances $\{\bfa X^{(i)},\bfa z^{(i)} \}_{i\in[n]}$ to be sampled i.i.d. from a distribution supported on $\Theta_{GM}$. Then, we can show the following generalization bound for the Transformer network in the space of $\Theta(B_{\bfa\theta},B_{M},B_{L})$.
\begin{proposition}[Generalization Bounds]\label{genbd}
    With probability at least $1-\delta$,
    \sm{\begin{align*}
        L&\lef(A_{\wha\theta}(\bfa H), \bfa P_1(\bfa z)\rig) \leq \inf_{\bfa\theta\in\Theta_{B_M,B_L}(B_{\bfa\theta})}\bb E[L\lef(A_{\bfa \theta}(\bfa H), \bfa P_1(\bfa z)\rig)]\\
        &+C\sqrt{\frac{D^2B_LB_M\log(NB_{\bfa\theta}B_{M}D\sigma m_0)+\log(2/\delta)}{n}},
    \end{align*}}
    where $m_0=\sup_{i\in[N],j\in[n]}\Vert\bfa\mu_i^{(j)}\Vert_2$.
\end{proposition}
\begin{remark}
    The above proposition implies that, given the sufficiently large number of samples $n$, the ERM solution generalizes to new samples as we can use the approximation results given by Theorem \ref{thm_main1} and \ref{thm_main2} to upper bound the first term on the R.H.S. of the inequality. The following results ultimately provide an ultimate bound for the error of the ERM estimator on the unseen instance.
\end{remark}
\begin{theorem}[The Matching Upper Bound]\label{ultimate}
    Let $r_k=\frac{\Delta}{\sigma}\sqrt{\frac{\alpha}{1+kd/N}}$, $k\log N = o(N\alpha^2)$, $M\asymp n^{1/2}$, $L\asymp k\log n$, $\sqrt k=o(r_k)$ as $n\to\infty$. Assume that we use Algorithm \ref{alg:example} as initialization and let $\tau > 4\log n+1$. Then, with probability at least $1-\delta - 5n^{-1} - 2\exp(-\Delta/\sigma)$, the ERM estimator given by the Transformer satisfies
    \begin{align*}
    L(&TF_{\wha\theta}(\bfa H),\bfa P_1(\bfa z))\lesssim \exp\lef(-(1+o(1))\frac{\Delta^2}{8\sigma^2}\rig)\\
    &+\sqrt kn^{-1/4}C^d\sqrt{Polylog(n)+\log(1/\delta)}+N^{-3/2}.
\end{align*}
And with the same parameter setup and initialization, with probability at least $1-\delta - 5n^{-1} - 2\exp(-\Delta/\sigma)$, the ERM estimator given by the Transformer+ satisfies
\begin{align*}
    L(&TF_{\wha\theta}^+(\bfa H),\bfa P_1(\bfa z))\lesssim \exp\lef(-(1+o(1))\frac{\Delta^2}{8\sigma^2}\rig)\\
    &+d\sqrt kn^{-1/4}C^d\sqrt{Polylog(n)+\log(1/\delta)}+N^{-100.5}.
\end{align*}
\end{theorem}
\begin{remark}
    Our results in the above theorem imply that, given the number of samples $n\asymp \exp\lef(\frac{\Delta^2}{2\sigma^2}\rig)$, Transformers can reach the fundamental limits given by Lemma \ref{minimaxlb} and achieve the minimax optimal rate of the clustering problem with high probability. Moreover, the introduction of the non-activated Attention layer in Softmax+ significantly improves the upper bound in the exponent of $N$. In particular, the $N^{-100.5}$ can even be improved with arbitrarily large universal constants. We discuss in section \ref{approx3} that solving a potential open problem on the universal approximation of the Transformer might lead to the removal of the non-activated Attention layer in the proof. 
\end{remark}
\subsection{The Proof Ideas}\label{approx3}
This section discusses some new results we obtained on the midway of proving Theorems \ref{thm_main1} and \ref{thm_main2}. We then present a proof sketch for the more complicated proof of Theorem \ref{thm_main1}.
\subsubsection{An Approximation Bound for the Softmax Function} We provide a new approximation bound for the sum of Softmax functions to mappings from $\bb R^{d_1}\to\bb R^{d_2}$. We first introduce the class of $(R,C_{\ell})$ smooth functions. The $(R, C_{\ell})$ smooth function class contains a wide range of functions.
\begin{definition}[\citep{bach2017breaking}\citep{bai2024transformers}]\label{rcsmoothfunc}
    A function $g:\bb R^d\to\bb R$ is $(R, C_{\ell})$ smooth if for $s= \lceil(k-1)/2\rceil+2$, $g$ is a $C^s$ function supported on $[-R, R]^k$ such that
    \begin{align*}
        \sup_{\bfa x\in [-R,R]^k}\Vert\nabla^ig(\bfa x)\Vert_{\infty}\leq L_i,
    \end{align*}
    for all $i\in\{0,1,\ldots,s\}$, with $\max_{0\leq i\leq s}L_iR^i\leq C_{\ell}$.
\end{definition}
Then, we are ready to present our results on the approximation error of Softmax functions.
 \begin{lemma}[Approximating $d$ Dimensional $(R, C_{\ell})$ Smooth Mappings by Softmax Neural Networks]\label{softmaxapprox}
        Consider an element-wise $(R, C_{\ell})$ smooth mapping $\bfa f(\bfa x)=(f_1(\bfa x),\ldots f_{d_1}(\bfa x))^\top$ where $\bfa x\in[-R,R]^d$. There exists a set of points $\{(\bfa A_{i}, a_{i})\}_{i\in[M]}$ with $\sup_{i\in[M]}\Vert\bfa A\Vert_2\leq C$ such that the following holds
        \begin{align*}
            \sup_{\bfa x\in\ca B\lef(\Vert\cdot\Vert_\infty, R\rig)}&\frac{1}{C_{\ell}}\Big\Vert\bfa f(\bfa x)-\sum_{i=1}^{Md} a_i\softmaxx\lef(\bfa A_i\begin{bmatrix}
                \bfa x\\
            1\end{bmatrix}\rig)\Big\Vert_{\infty}\\
            &\leq C(f)^d\sqrt{\frac{dd_1}{M}\log\lef(\frac{MR}{dd_1}\rig)}.
        \end{align*}
 \end{lemma}
\begin{remark}
    The above bound demonstrates the universal approximation of Softmax functions to smooth mappings. Our proof idea utilizes a preliminary result on the sigmoid function and dissects the softmax function into multiple sigmoid functions. For each of the sigmoid functions, we use the probabilistic method to construct $L_{\infty}$ approximation bound.
\end{remark}
In Lemma \ref{softmaxapprox}, our proof applies to the left product of $\bfa A$. It is then of general interest to know whether there exists a universal approximation bound for the right product form $\softmax\lef(\begin{bmatrix}
        \bfa x&1
\end{bmatrix}\bfa A\rig)$. Solving this fundamental problem helps us to achieve the rate of the Transformer+ using the Transformer model.
\subsubsection{The Proof Sketches of Theorem \ref{thm_main1}}
We here provide the proof sketch of the Theorem \ref{thm_main1}. The proof of Theorem \ref{thm_main2} is more involved in the Expectation step and is delayed to the appendix. Our proof idea is to manually construct $\bfa\theta$ for the network and estimate the error caused by each layer constructed. 

\paragraph{The Expectation Step.} In the expectation step, we notice the following relationship holds
\sm{\begin{align*}
   \begin{bmatrix}
       \bfa X_1&\ldots&\bfa X_N
   \end{bmatrix} \softmax\lef(\begin{bmatrix}
        \bfa p_{1,1}^{(0)}\\\vdots\\\bfa p_{1,N}^{(0)}
    \end{bmatrix}\rig)\approx \begin{bmatrix}
        \wha\mu_1^{(1)}&\ldots&\wha\mu_k^{(1)}
    \end{bmatrix}.
\end{align*}}
However, we have \tny{$\begin{bmatrix}
    \bfa p_{1,1}^{(0)}&\ldots &\bfa p_{1,N}^{(0)}\\
    \bfa 0&\ldots&\bfa 0
\end{bmatrix}^\top$} and the $\bfa 0$ part needs to be cancelled. We then construct another head with $\bfa 0$ matrix in the SoftMax function to cancel out the $\bfa 0$ part in the first head. The two cancellations result in an approximation error of $O(1/N)$. 

\paragraph{The Maximization Step} In the maximization step, our proof involves a total of 4 steps. Our initial matrix is given by 
\tny{\begin{align*}
    \begin{bmatrix}
        \bfa X_1&\bfa X_2&\ldots&\bfa X_{k}&\ldots&\bfa X_N\\
        \wha \mu_{1}^{(1)}&\wha \mu_{2}^{(1)}&\ldots&\wha \mu_{k}^{(1)}&\ldots&\bfa 0\\
        \bfa p^{(0)}_{1,1}&\bfa p^{(0)}_{1,2}&\ldots&\bfa p_{1,k}^{(0)}&\ldots&\bfa p^{(0)}_{1,N}\\
        \bfa p_{2,1}&\bfa p_{2,2}&\ldots&\bfa p_{2,k}&\ldots&\bfa p_{2,N}\\
        1&1&\ldots&1&\ldots&1\\
        &&\bfa 0&&&
    \end{bmatrix}.
\end{align*}}
Then, in the \textbf{Step 1}, we copy in a total of $k$ times the first row and move them to the $\bfa 0$ part using two FC layers, with one moving the negative part and one moving the positive part, providing
\tny{\begin{align*}
    \begin{bmatrix}
        \vdots&\vdots&\vdots&\vdots&\vdots&\vdots\\
        1&1&\ldots&1&\ldots&1\\
        \bfa X_{1,1}&\bfa X_{2,1}&\ldots&\bfa X_{k,1}&\ldots&\bfa X_{N,1}\\
        &&\vdots\\
        \bfa X_{1,k}&\bfa X_{2,k}&\ldots&\bfa X_{k,k}&\ldots&\bfa X_{N,k}\\
        &&\bfa 0
    \end{bmatrix}.
\end{align*}}
Then, in the \textbf{Step 2}, we move $\{\wha\mu_i^{(1)}\}_{i\in[M]}$ to $\{\bfa x_{j,i}\}_{j\in[N]}$, yielding
\tny{$$\begin{bmatrix}
        \vdots&\vdots&\vdots&\vdots&\vdots\\
        1&\ldots&1&\ldots&1\\
        \bfa X_{1,1}-\wha\mu_1^{(1)}&\ldots&\bfa X_{k,1}-\wha\mu_1^{(1)}&\ldots&\bfa X_{N,1}-\wha\mu_1^{(1)}\\
        &&\vdots\\
        \bfa X_{1,k}-\wha\mu_k^{(1)}&\ldots&\bfa X_{k,k}-\wha\mu_k^{(1)}&\ldots&\bfa X_{N,k}-\wha\mu_k^{(1)}\\
        &&\bfa 0
    \end{bmatrix},$$}
    This step utilizes the approximation bound given by Lemma \ref{softmaxapprox} to approximate the function of $f(\bfa x)=\bfa x_i$.
    
    Then, in the \textbf{Step 3}, we apply the approximation bound again to construct Softmax networks that approximate the mapping from a vector to its norm, providing us with the following matrix.
    \ttny{$$\begin{bmatrix}
        \vdots&\vdots&\vdots&\vdots&\vdots\\
        1&\ldots&1&\ldots&1\\
        \Vert\bfa X_{1,1}-\wha\mu_1^{(1)}\Vert_2&\ldots&\Vert\bfa X_{k,1}-\wha\mu_1^{(1)}\Vert_2&\ldots&\Vert\bfa X_{N,1}-\wha\mu_1^{(1)}\Vert_2\\
        &&\vdots\\
        \Vert\bfa X_{1,k}-\wha\mu_k^{(1)}\Vert_2&\ldots&\Vert\bfa X_{k,k}-\wha\mu_k^{(1)}\Vert_2&\ldots&\Vert\bfa X_{N,k}-\wha\mu_k^{(1)}\Vert_2\\
        &&\bfa 0
    \end{bmatrix},$$}
    Finally, in the \textbf{Step 4}, we obtain approximate vectors to $\{\bfa p_{1,i}^{(1)}\}_{i\in[N]}$ through applying the softmax function to the submatrix
    \tny{\begin{align*}
        \begin{bmatrix}
        \Vert\bfa X_{1,1}-\wha\mu_1^{(1)}\Vert_2&\ldots&\Vert\bfa X_{N,1}-\wha\mu_1^{(1)}\Vert_2\\
        \vdots&\vdots&\vdots\\
        \Vert\bfa X_{1,k}-\wha\mu_1^{(1)}\Vert_2&\ldots&\Vert\bfa X_{N,k}-\wha\mu_1^{(1)}\Vert_2
    \end{bmatrix}.
    \end{align*}} Using another approximation bound showing the difference between the Softmax and the Hardmax function we accomplish the Maximization step.

\section{Simulations}\label{sect4}

\begin{figure}[!h]
% \vspace{-1em}
\minipage{0.48\textwidth}
    \minipage{0.48\textwidth}
        \includegraphics[width=\linewidth]{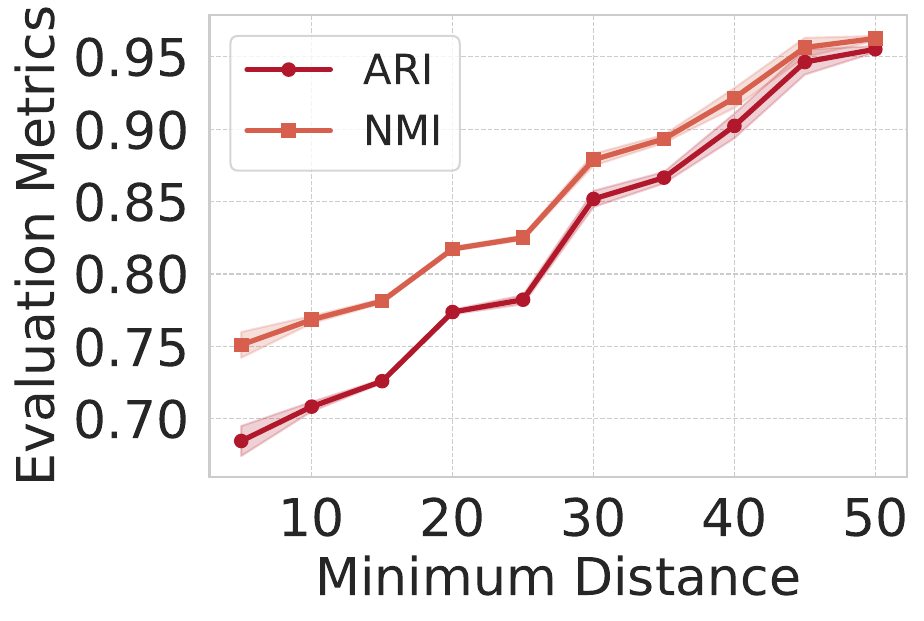}
    \endminipage\hfill
    \minipage{0.48\textwidth}
        \includegraphics[width=\textwidth]{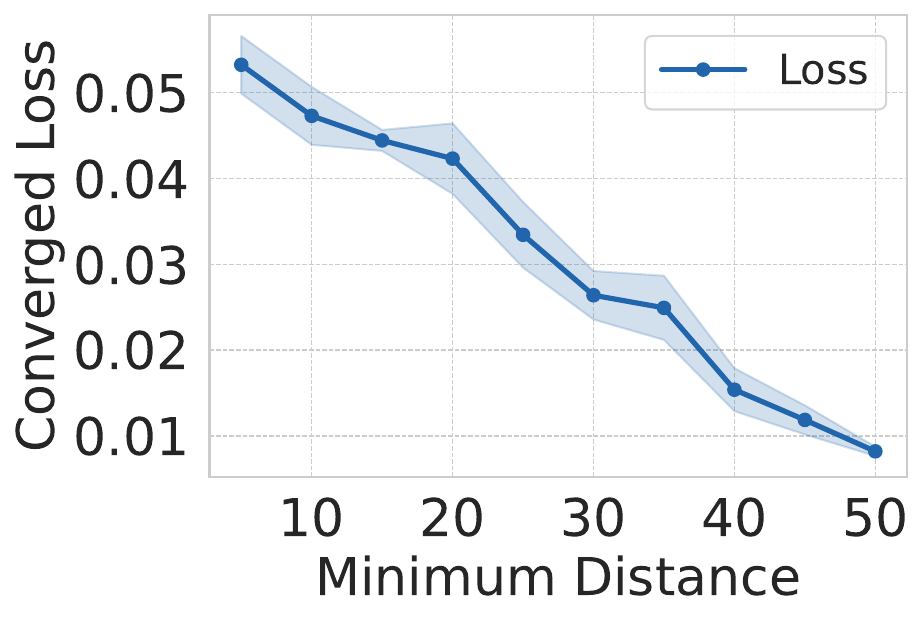}
    \endminipage\hfill
\endminipage\hfill
\minipage{0.48\textwidth}
    \minipage{0.48\textwidth}
        \includegraphics[width=\textwidth]{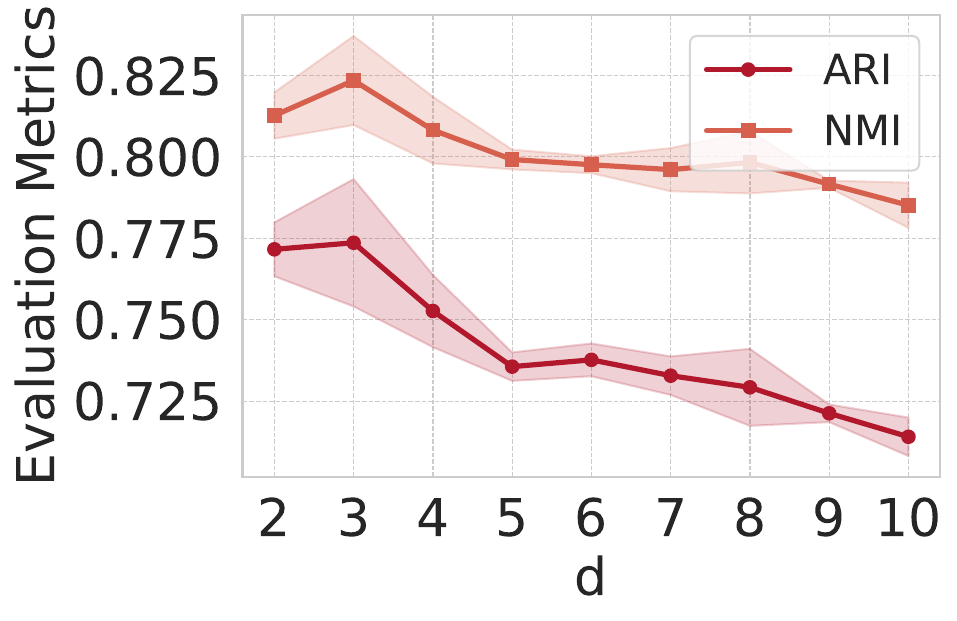}
    \endminipage\hfill
    \minipage{0.48\textwidth}
        \includegraphics[width=\textwidth]{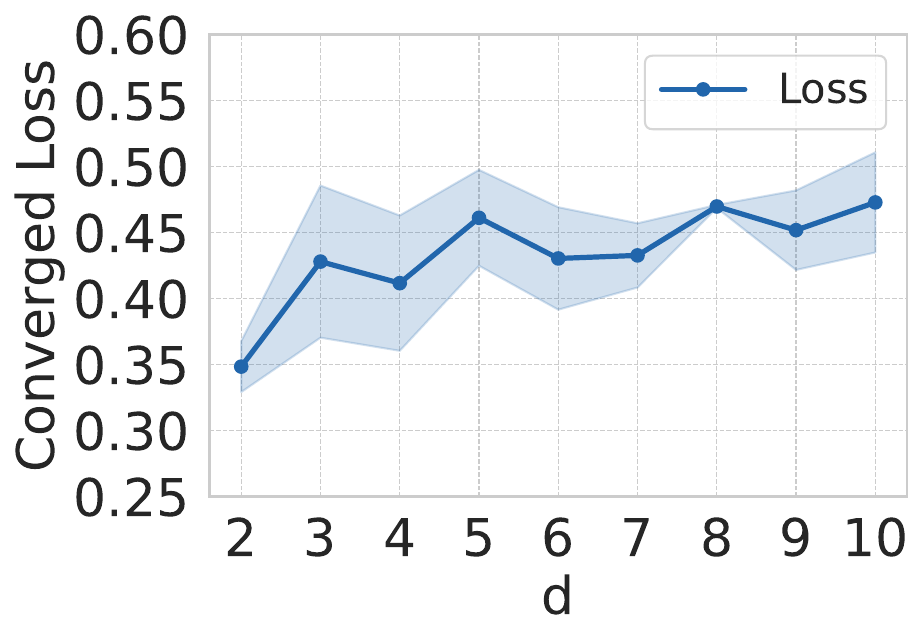}
    \endminipage\hfill
\endminipage\hfill
\minipage{0.48\textwidth}
    \minipage{0.48\textwidth}
        \includegraphics[width=\textwidth]{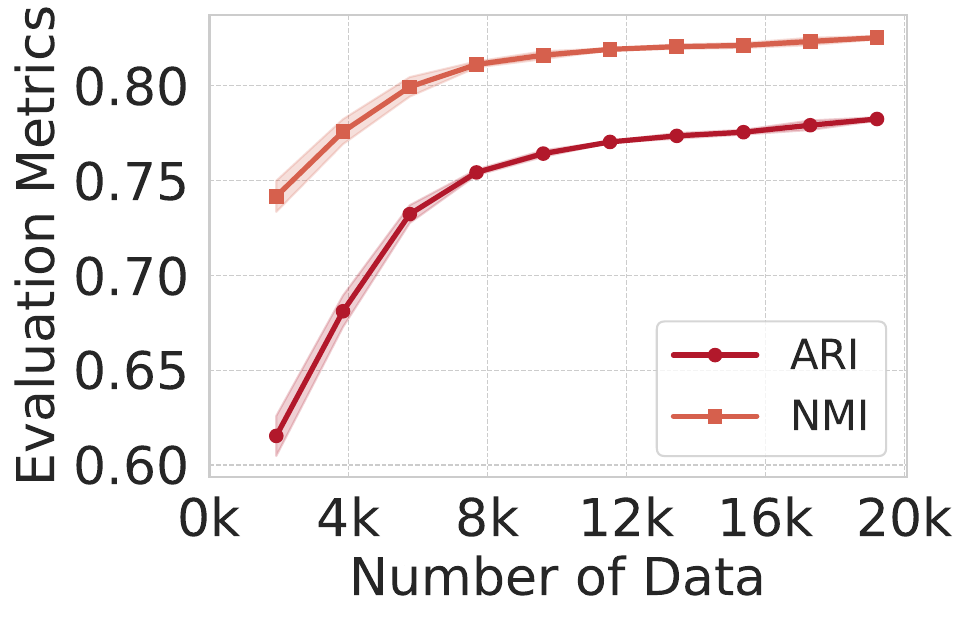}
    \endminipage\hfill
    \minipage{0.48\textwidth}
        \includegraphics[width=\textwidth]{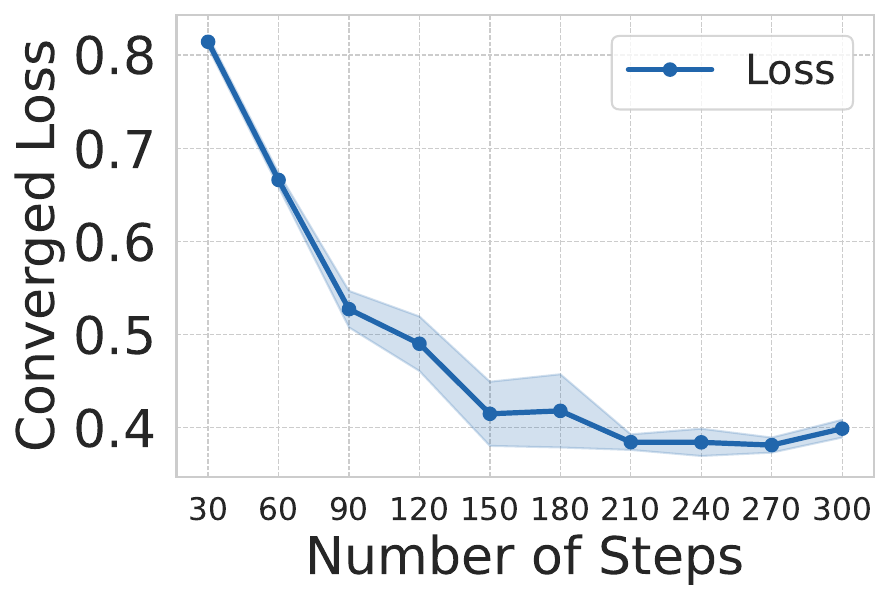}
    \endminipage\hfill
\endminipage
\vspace{-1em}
\caption{\textbf{$4$-Class Clustering with Different Minimum Distance, Data Dimension, and Number of Training Data.} 
% \red{Can move detials into appendix.}
We train a small Transformer (layer $=3$, head $=2$, embedding $= 64$) and iterate for $300$ steps for each different setting. Each point in the figure is evaluated on $512$ testing data. 
We report the $10$ runs averaged result with a shaded region representing the standard deviation.
Each training sample is generated according to isotropic Gaussian with covariances $\sigma^2 \bfa I$.
\emph{(1) First Row: Minimum Distance.} We set $\sigma^2 \sim \mathrm{Uniform}[10,40]$.
\emph{(2) Second Row: Data Dimension.} We set $\sigma^2 \sim \mathrm{Uniform}[10,20]$, minimum distance $=5$.
\emph{(3) Three Row: Number of Training Data.} 
We set $\sigma^2 \sim \mathrm{Uniform}[0.5,5]$, minimum distance $=5$.}
\label{fig:dist_dim_step}
% \vspace{-1em}
\end{figure}

\begin{figure}[!h]
% \vspace{-1em}
\minipage{0.48\textwidth}
    \minipage{0.48\textwidth}
        \includegraphics[width=\linewidth]{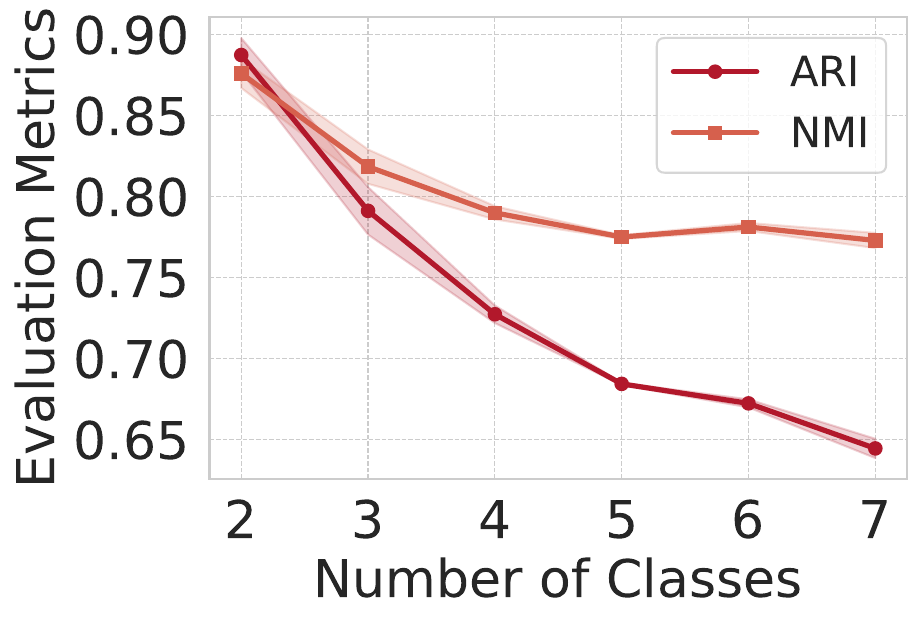}
    \endminipage\hfill
    \minipage{0.48\textwidth}
        \includegraphics[width=\textwidth]{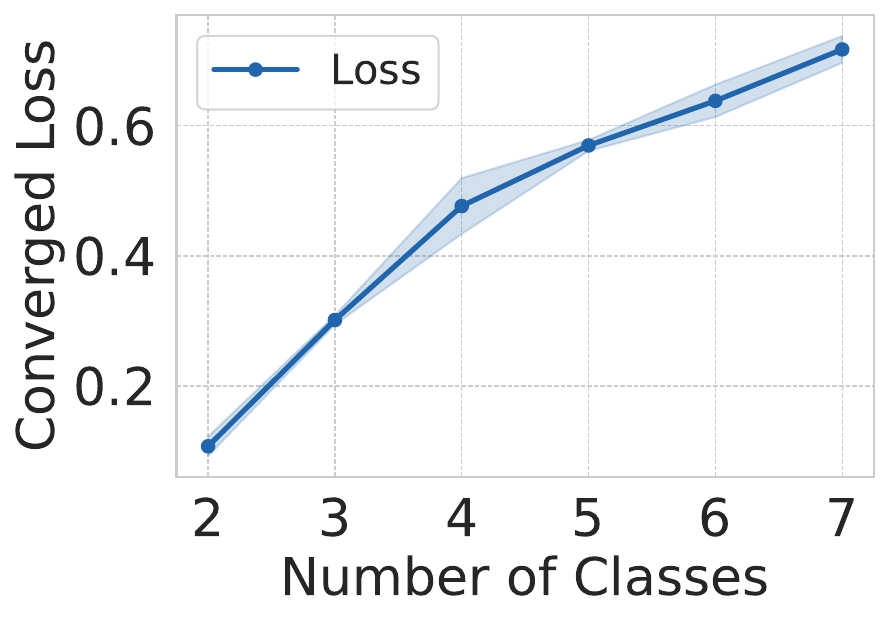}
    \endminipage\hfill
\endminipage\hfill
\minipage{0.48\textwidth}
    \minipage{0.48\textwidth}
        \includegraphics[width=\textwidth]{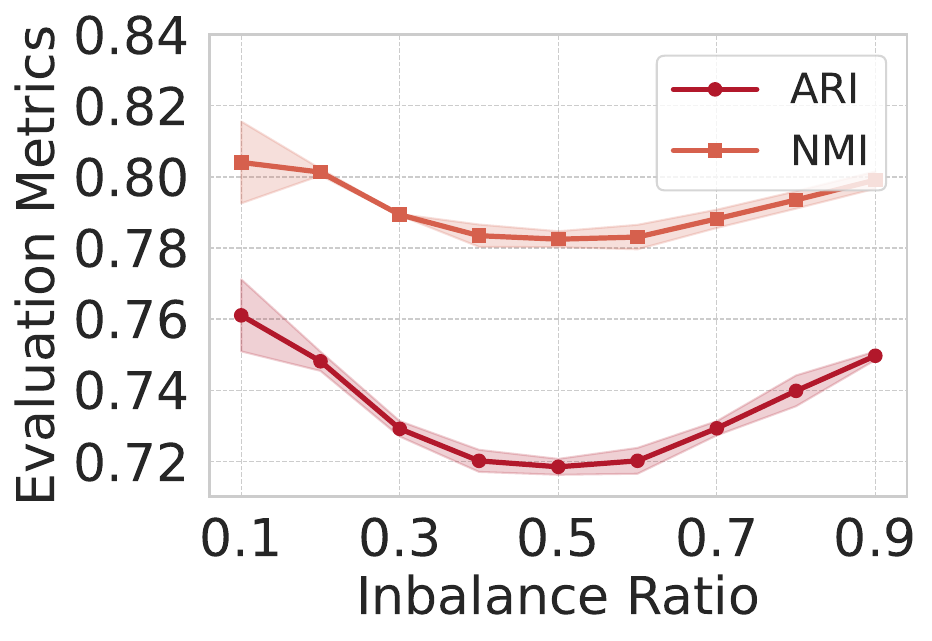}
    \endminipage\hfill
    \minipage{0.48\textwidth}
        \includegraphics[width=\textwidth]{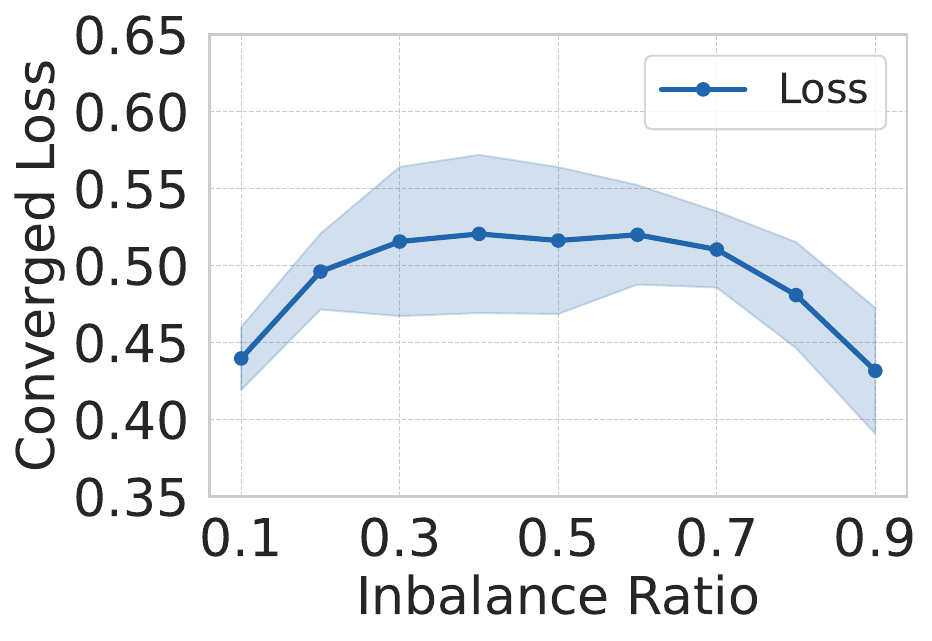}
    \endminipage\hfill
\endminipage
\vspace{-1em}
\caption{\textbf{$4$-Class Clustering with Different Number of Class and Inbalance Ratio.}
% \red{Can move detials into appendix.}
We train a small Transformer (layer $=3$, head $=2$, embedding $= 64$) and train for $300$ steps for each different setting. Each point in the figure is evaluated on $512$ testing data.
We report the $10$ runs averaged result with a shaded region representing the standard deviation.
Each training sample is generated according to isotropic Gaussian with covariances $\sigma^2 \bfa I$.
\emph{(1) First Row: Number of Class.} We set $\sigma^2 \sim \mathrm{Uniform}[10,20]$, minimum distance $=5$.
\emph{(2) Second Row: Inbalance Ratio.} Two clusters each contain 50 data points, while the other two contain $50 \times \mathrm{ratio}$ and $50 \times \mathrm{1-ratio}$ respectively. We set $\sigma^2 \sim \mathrm{Uniform}[10,20]$, minimum distance $=5$.}
\label{fig:inbalance}
% \vspace{-1em}
\end{figure}

% We set $\sigma^2 \sim \mathrm{Uniform}[10,20]$, minimum distance $=5$.}

\begin{figure}[!h]
% \vspace{-1em}
\minipage{0.48\textwidth}
    \minipage{0.48\textwidth}
        \includegraphics[width=\textwidth]{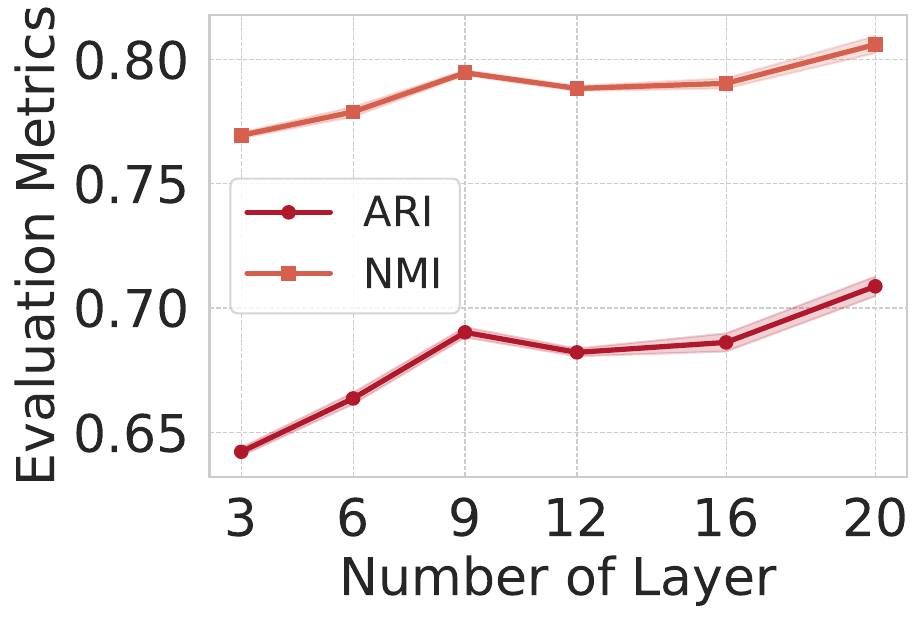}
    \endminipage\hfill
    \minipage{0.48\textwidth}
        \includegraphics[width=\textwidth]{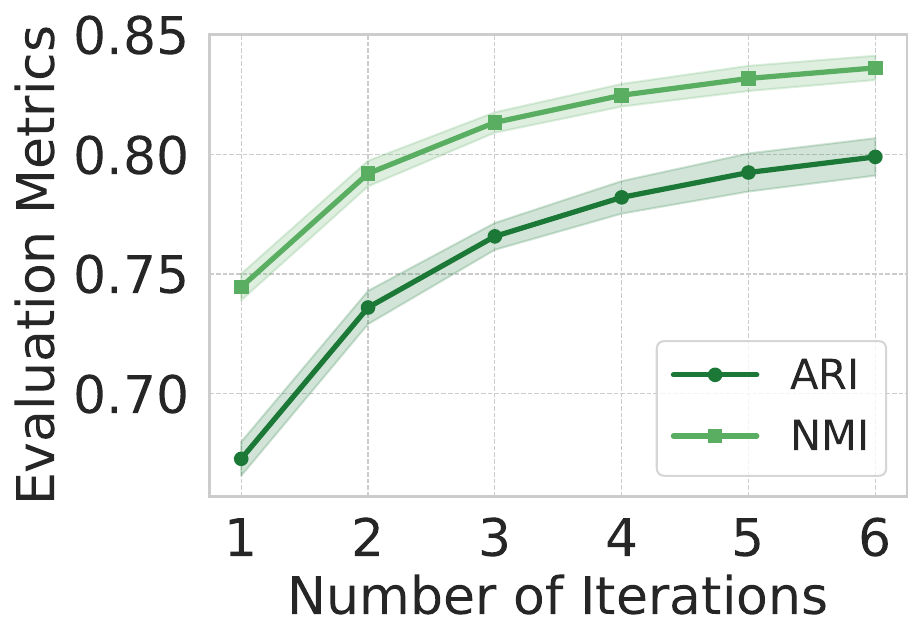}
    \endminipage\hfill
\endminipage
\vspace{-1em}
\caption{\textbf{Comparision between Transformer and Lloyd's Algorithm.}
% \red{Can move detials into appendix.}
We compare the effect of the number of layers in Transformers with the number of iterations $\tau$ in Lloyd's algorithm under the same dataset configuration.  
We use a $6$-class dataset, where each cluster contains $50$ data points in a $d=10$ dimensional space.  
Each training sample is generated according to isotropic Gaussian with covariances $\sigma^2 \bfa I$, where $\sigma^2 \sim \mathrm{Uniform}[20,30]$, and the minimum cluster separation is set to $1$.  
\emph{(1) Left: Transformer.} 
We train Transformers with fixed head $=2$, embedding $= 64$, but vary the number of layers from $3$ to $20$.
Each model is trained for $500$ steps per layer.  
\emph{(2) Right: Lloyd's Algorithm.}
% We use ``sklearn`` to run the Lloyd algorithm, and vary the maximum iteration from $1$ to $6$.
We use \texttt{sklearn}\citep{pedregosa2011scikit} to run the Lloyd's algorithm, varying the maximum iteration count from $1$ to $6$.  
Early convergence is declared when the Frobenius norm of the difference between cluster centers in consecutive iterations falls below $10^{-4}$. 
Each point in the figure represents an evaluation of $512$ test samples.  
Results are averaged over $10$ runs, with the shaded region indicating the standard deviation.}
\label{fig:com_lloyd}
% \vspace{-1em}
\end{figure}

% \red{TODOs: Comparision of Lloyd baseline and Transformer.}
In this section, we verify our theoretical results on the multi-class clustering problem and examine its interplay with five key factors: the minimum distance between centroids $\Delta$, the data dimension $d$, the training sample size $N$, the total number of classes, and an imbalance ratio $\alpha$. These results are presented in  \cref{fig:dist_dim_step} and \cref{fig:inbalance}.
Furthermore, we compare the impact of a number of layers in the Transformer with the number of iterations $\tau$ in \cref{fig:com_lloyd}.

\paragraph{Experimental Setup}
We use a small Transformer with $3$ layers, $2$ heads, and $64$-dimensional embedding size.
All simulations are conducted on NVIDIA A100 80G GPUs. We run each experiment for 300 iterations, initialize the model with $10$ different random seeds, and report the mean and standard deviation of the resulting metrics. The model is trained using the Adam optimizer with a learning rate of $0.0005$ and an exponential decay factor of $0.995$ for each step.
After training, each configuration is evaluated on 512 synthetic and random test samples. Note that our empirical evaluation slightly differs from the theoretical part through removing the auxiliary matrix $\bfa P$ given by \eqref{aux} from the input.

\paragraph{Metrics.}
We compute cross entropy among every permutation of the label and choose the minimum as the loss function since clustering tasks are permutation invariant.
We evaluate the clustering performance using two widely adopted permutation-invariant metrics: Adjusted Rand Index (ARI) and Normalized Mutual Information (NMI) \citep{ma2019learning, huang2020partially, monnier2020deep, sun2024lsenet, li2024image}.

\paragraph{Preparation for the Synthetic Data.}
{We generate our synthetic data as follows: For each input $\bfa X \in \R^{d \times N}$, we sample $50$ data points from every cluster. 
Each sample is generated according to isotropic Gaussian with covariances $\sigma^2 \bfa I$.
The variance $\sigma$ differs from task to task; we specify more details in the caption of figures.}

\paragraph{Results}
Our results suggest that the theoretical threshold given by the minimax rate matches with the trend given in the experiments. Moreover, we also showcase that the pre-trained Transformers can be a strong alternative to Lloyd's algorithm, verifying the strong inference capacities of Transformers on this problem.

\section{Discussions}\label{sect5}
This section discusses the limitations on the theory part of this work and points to future working directions. 
\paragraph{Limitations.} Our limitations in the theoretical results can be summarized as follows: \textbf{(1)} From the theoretical perspective, our results guarantee the performance of ERM solutions whereas the true estimator is obtained through stochastic gradient descent method; \textbf{(2)} Our theoretical results utilize the context-augmented matrix $\bfa P$, which is verified removable from our empirical results. 
\paragraph{Future Works.} Beyond resolving the limitations in this work, other future working directions from this work include: \textbf{(1)} Taking into consideration of the layer norm in the Transformer architecture. \textbf{(2)} Resolving the universal approximation problem raised in section \ref{approx3}; \textbf{(3)} Removing the initialization procedure in the theory.

\section*{Impact Statement}
This paper theoretically analyzes the capability of Transformers in performing EM algorithm. Due to the theoretical nature of this work, there is no negative sociatal impact.
% Due to the theoretical nature of this work, the sociatal impact is minor. Though it is possible that Transformer models might induce safety issues when used improperly.

\bibliography{bib.bib}
\bibliographystyle{icml2025}

\newpage
\appendix
\onecolumn
\section{Theoretical Background}
This section provides approximation results of the Softmax function in the form of $f(\bfa x):=\sum_{i=1}^M a_i\sigmoid\lef(\bfa x^\top\bfa v_i\rig)+a_0$ where $\{a_i\}_{i\in[M]}\subset \bb R$ and $\{\bfa v_i\}\subset \bb R^{d+1}$. We consider the class of $(R, C_{\ell})$-smooth functions \citep{bach2017breaking,bai2024transformers} defined as follows.
\begin{definition}[\citet{bai2024transformers}]\label{rcsmoothfunc}
    A function $g:\bb R^d\to\bb R$ is $(R, C_{\ell})$ smooth if for $s= \lceil(k-1)/2\rceil+2$, $g$ is a $C^s$ function supported on $[-R, R]^k$ such that
    \begin{align*}
        \sup_{\bfa x\in [-R,R]^k}\Vert\nabla^ig(\bfa x)\Vert_{\infty}\leq L_i,
    \end{align*}
    for all $i\in\{0,1,\ldots,s\}$, with $\max_{0\leq i\leq s}L_iR^i\leq C_{\ell}$.
\end{definition}
We then consider the following class of functions
\begin{align*}
    &\ca F_d = \bigg\{f:f(\bfa x)=\int_{\ca W:\Vert\bfa v\Vert=1}\sigmoid\lef([\bfa x^\top,1]\bfa v\rig)d\mu(\bfa v)\bigg\},\text{ with } TV(\mu)=\int_{\ca W:\Vert\bfa v\Vert_2=1}d|\mu(\bfa v)|<C_{\ell}C(f)^d ,
\end{align*}
where $C(f)$ is a constant that depends only on $f$. 
It is further noted that by \citet{bach2017breaking}, we can write that the class of $(R, C_{\ell})$ smooth functions belongs to the above class. Then we prove the following approximation lemma for the sigmoid function, which provides explicit dependence on $B$ and $C$.
 \begin{lemma}\label{sigmoidapprox}
     Suppose $f$ is $(R,C_{\ell})$ smooth. Then there exists a set of points $(\bfa v_1, a_1),\ldots,(\bfa v_M, a_M)\in \ca B(\Vert\cdot\Vert_2, 1)$ that makes the following hold
     \begin{align*}
         \sup_{\bfa x\in \ca B\lef(\Vert\cdot\Vert_\infty,R\rig)}\Big|\frac{1}{C_{\ell}}f(\bfa x)-\frac{1}{M}\sum_{i=1}^Ma_i\sigmoid\lef([\bfa x^\top,1]^\top\bfa v_i\rig)\Big|&\leq  C(f)^d\inf_{\epsilon>0}\Big(2\epsilon+\sqrt{\frac{\log(\ca N(\ca B(\Vert\cdot\Vert_{\infty},R),\epsilon)/\delta)}{M}}\Big)\\
         &\lesssim C(f)^d\sqrt{\frac{d}{M}\log\Big(\frac{MR}{d}\Big)},
     \end{align*}
     where we also have $\sum_{i=1}^M|a_i|\leq C(f)^d$.
 \end{lemma}
 \begin{proof}
     The proof goes by the probabilistic method, where we can first use \citet{pisier1981remarques} to show that when we sample from the distribution given by $f(\bfa v)=\frac{|\mu(\bfa v)|}{\int_{\bb S^d}d\mu(\bfa v)}=\frac{|\mu(\bfa v)|}{TV(\mu)}$. Then, under this probability measure, we sample independently in a total of $M$ samples $(\bfa V_1, \ldots, \bfa V_M)$ to obtain that for any $\bfa x\in\ca V$, pointwise
     \begin{align*}
         \int_{\bb R^{d+1}}&\softmaxx\lef([\bfa x^\top, 1]\bfa v\rig)d\mu(\bfa v)- \frac{TV(\mu)}{M}\sum_{i=1}^M\softmaxx\lef([\bfa x^\top, 1]\bfa V_i\rig)\\
         &= TV(\mu)\bl\bb E\Big[\sign(\mu(\bfa v)) \softmaxx\lef([\bfa x^\top,1]\bfa V\rig)\Big]-\frac{1}{M}\sum_{i=1}^M\sign\lef(\mu(\bfa v)\rig)\softmaxx\lef([\bfa x^\top,1]\bfa V_i\rig)\br.
     \end{align*}
     And we have by Hoeffding's inequality,
     \begin{align*}
         \bb P\bl\Big|\bb E\lef[\sign(\mu(\bfa v))\softmaxx\lef([\bfa x^\top, 1]\bfa V\rig)\rig]-\frac{1}{M}\sum_{i=1}^M\sign(\sigma_f(\bfa V_i))\softmaxx\lef([\bfa x^\top, 1]\bfa V_i\rig)\Big|\geq \frac{t}{TV(\mu)}\br\leq\exp\lef(-\frac{CMt^2}{TV(\mu)^2}\rig).
     \end{align*}
     And, by the union bound, we can show that for points in the $\epsilon$-cover of $\ca B(\Vert\cdot\Vert_{\infty},R)$, the following holds
     \begin{align*}
         \bb P\bl\sup_{\bfa x\in \ca N(\ca B(\Vert\cdot\Vert_{\infty},R),\epsilon)}&\Big|\bb E\Big[\ub{\sign(\sigma_f (\bfa V))\softmaxx\lef([\bfa x^\top, 1]\bfa V\rig)}_{=:F(\bfa V,\bfa x)}\Big]-\frac{1}{M}\sum_{i=1}^M\sign(\sigma_f(\bfa V_i))\softmaxx\lef([\bfa x^\top,1]\bfa V_i\rig)\Big|\geq\frac{t}{TV(\mu)}\br\\
         &\leq |\ca N(\ca B(\Vert\cdot\Vert_{\infty},R),\epsilon)|\exp\lef(-\frac{CMt^2}{TV(\mu)^2}\rig).
     \end{align*}
     Alternatively, we can show the following holds with probability at least $1-\delta$,
     \begin{align*}
         \sup_{\bfa x\in\ca N(\ca B(\Vert\cdot\Vert_{\infty},R),\epsilon)}&\Big|\bb E\lef[\sign(\mu(\bfa v))\softmaxx\lef([\bfa x^\top,1]\bfa V\rig)\rig]-\frac{1}{M}\sum_{i=1}^M\sign(\sigma_f)\softmaxx\lef([\bfa x^\top,1]\bfa V_i\rig)\Big|\\
         &\leq \sqrt{\frac{\log\lef(\ca N(\ca B(\Vert\cdot\Vert_{\infty},R),\epsilon)/\delta\rig)}{M}}.
     \end{align*}

     Then we consider generalizing these results to uniform convergence. For $\bfa x$, we denote $\pi(\bfa x)$ as the closest point in the $\epsilon$-cover of $\ca B(\Vert\cdot\Vert_{\infty},R)$ denoted by $\ca N(\ca B(\Vert\cdot\Vert_{\infty},R),\epsilon)$. For function $F$, we can show that for all $\bfa x\in\ca B(\Vert\cdot\Vert_{\infty},R)$,
     \begin{align*}
         \sup_{\bfa x\in\ca B(\Vert\cdot\Vert_{\infty},R)}\lef|\bb E[F(\bfa V,\bfa x)] - \bb E[F(\bfa V,\pi(\bfa x))]\rig|\leq \lef|\bfa V^\top(\bfa X_1-\bfa X_2)\rig|\leq\Vert\bb E[\bfa V^\top]\Vert_2\Vert\bfa X_1-\bfa X_2\Vert_2\leq \epsilon.
     \end{align*}
     Then we consider the error given by
     \begin{align*}
         \sup_{\bfa x\in\ca B(\Vert\cdot\Vert_{\infty},R)}\Big|\frac{1}{M}\sum_{i=1}^MF(\bfa V_i,\bfa x)-\frac{1}{M}\sum_{i=1}^MF(\bfa V_i,\pi(\bfa x))\Big|\leq\frac{1}{M}\sum_{i=1}^M\epsilon\Vert\bfa V_i\Vert_2\leq\epsilon.
     \end{align*}
    Then the following holds with probability at least $1-\delta$,
     \begin{align*}
         &\sup_{\bfa x\in\ca B(\Vert\cdot\Vert_{\infty},R)}\Big|\bb E\lef[F(\bfa V, \bfa x)\rig]-\frac{1}{M}\sum_{i=1}^MF(\bfa V_i,\bfa x)\Big|\\
         &\leq \sup_{\bfa x\in\ca B(\Vert\cdot\Vert_{\infty},R)}\Big|\bb E\lef[F(\bfa V, \bfa x)\rig]-\bb E\lef[F(\bfa V, \pi(\bfa x))\rig]\Big|+\sup_{\bfa x\in\ca B(\Vert\cdot\Vert_{\infty},R)}\Big|\frac{1}{M}\sum_{i=1}^MF(\bfa V_i,\bfa x)-\frac{1}{M}\sum_{i=1}^MF(\bfa V_i, \pi(\bfa x))\Big|\\
         &+\sup_{\pi(\bfa x)\in\ca N(\ca B(\Vert\cdot\Vert_{\infty},R),\epsilon)}\Big|\bb E\lef[F(\bfa V,\pi(\bfa x))\rig]-\frac{1}{M}\sum_{i=1}^MF(\bfa V_i,\pi(\bfa x))\Big|\leq\epsilon+\epsilon + \sqrt{\frac{\log(\ca N(\ca B(\Vert\cdot\Vert_{\infty},R),\epsilon)/\delta)}{M}}.
     \end{align*}
     Given these results, we show that there exists a set of parameters $\{(\bfa V_i, a_i=TV(\mu)\sign_f(\bfa V_i))\}_{i\in[m]}$ where the following holds
     \begin{align*}
         \sup_{\bfa x\in\ca B(\Vert\cdot\Vert_{\infty},R)}\Big|\frac{1}{C_{\ell}}f(\bfa x)-\frac{1}{M}\sum_{i=1}^M a_i\softmaxx\lef([\bfa x^\top, 1]\bfa V_i\rig)\Big|&\lesssim \inf_{\epsilon}\bigg\{\epsilon+\sqrt{\frac{\log(\ca N(\ca B(\Vert\cdot\Vert_{\infty},R),\epsilon))}{M}}\bigg\}\\
         &\lesssim \inf_{\epsilon}\bigg\{\epsilon +\sqrt{\frac{d\log\frac{R}{\epsilon}}{M}}\bigg\}\lesssim\sqrt{\frac{d}{M}\log\lef(\frac{MR}{d}\rig)},
     \end{align*}
     where we already utilize the estimate given by \citet{wu2020information} on the covering number of $L_2$ balls.
 \end{proof}
 \begin{lemma}[Approximating $d$ Dimensional $(R, C_{\ell})$ smooth functions by Softmax Neural Networks]\label{softmaxapprox}
        Consider an element-wise $(R, C_{\ell})$ smooth mapping $\bfa f(\bfa x)=(f_1(\bfa x),\ldots f_k(\bfa x))^\top$ where $\bfa x\in[-R,R]^d$. There exists a set of points $\{(\bfa A_{i}, a_{i})\}_{i\in[M]}$ with $\sup_{i\in[M]}\Vert\bfa A\Vert_2\leq C$ such that the following holds
        \begin{align*}
            \sup_{\bfa x\in\ca B\lef(\Vert\cdot\Vert_\infty, R\rig)}\frac{1}{C_{\ell}}\Big\Vert\bfa f(\bfa x)-\sum_{i=1}^{Md} a_i\softmaxx\lef(\bfa A_i\begin{bmatrix}
                \bfa x\\
1\end{bmatrix}\rig)\Big\Vert_{\infty}\leq C(f)^d\sqrt{\frac{d^2}{M}\log\lef(\frac{MR}{d^2}\rig)}.
        \end{align*}
 \end{lemma}
 \begin{proof}
     Our proof goes by connecting the $\softmax$ activation with the $\sigmoid$ functions. Note that by lemma \ref{sigmoidapprox} we can show that for all $\ell\in[k]$, there exists a set $\{(\bfa v_i^{(\ell)}, a_i^{(\ell)})\}_{\ell\in[M^\prime]}$
     \begin{align*}
         \sup_{\bfa x\in\ca B\lef(\Vert\cdot\Vert_{\infty}, R\rig)}\frac{1}{C_{\ell}}\Big|f_{\ell}(\bfa x)-\sum_{i=1}^{M^\prime}a_i^{(\ell)}\sigmoid\lef(\bfa v_i^{(\ell),\top}\begin{bmatrix}
             \bfa x\\
             1
         \end{bmatrix}\rig)\Big|\leq C(f)^d\sqrt{\frac{d}{M^\prime}\log\lef(\frac{M^\prime R}{d}\rig)}.
     \end{align*}
     Consider the following matrices construction of $\{\bfa B_{i}^{(\ell)}\}_{i\in[d], \ell\in[M]}\subset\bb R^{k\times d}$, given by 
            $\bfa B_{i}^{(\ell)} = \begin{bmatrix}
            \bfa 0_{(\ell-1)\times (d-1)}& 0\\
             \bfa v_{i,[d-1]}^{(\ell),\top}&\log d+\bfa v_{i,d}^{(\ell)}\\
             \bfa 0&\bfa 0
         \end{bmatrix}$. Then we can show that
         \begin{align*}
             \sup_{\bfa x\in\ca B(\Vert\cdot\Vert_{\infty})}\bigg\Vert\begin{bmatrix}
                 \bfa 0_{(\ell-1)\times 1}\\
                 f_{\ell}(\bfa x)\\
                 \bfa 0
             \end{bmatrix}-\sum_{i=1}^{M^\prime}a_i^{(\ell)}\softmaxx\lef(\bfa B_i^{(\ell)}\begin{bmatrix}
                 \bfa x\\
                 1
             \end{bmatrix}\rig)\bigg\Vert_{\infty}\leq C(f)^d\sqrt{\frac{d}{M^\prime}\log\Big(\frac{M^\prime R}{d}\Big)},
         \end{align*}
         which completes the proof through noticing that $M^\prime d=M$.
 \end{proof}
    \begin{lemma}[Norm Approximation by Sigmoid Functions]\label{Softmaxapprox1/x}
    Consider the vector $\bfa v\in\bb R^d$. Assume that there exists a constant $C$ with $\Vert\bfa v\Vert_2\leq C$. For $M<\lef(C\frac{\overline R}{\underline R}\rig)^d\frac{1}{\epsilon^2}\log(1+C/\epsilon)$ such that there exists $\{\bfa a_m\}_{m\in[M]}\subset\bb S^{d}$ and $\{c_m\}_{m\in[M]}\subset\bb R$ where for all $\bfa v$ with $\overline R\geq\Vert\bfa v\Vert_2\geq \underline R$, we have 
    \begin{align*}
        \bigg|\sum_{m=1}^Mc_m\sigmoid\lef(\bfa a_m^\top\begin{bmatrix}
            \bfa v\\
            1
        \end{bmatrix}\rig)-\frac{1}{\Vert\bfa v\Vert_2}\bigg|\leq \lef(\frac{C\overline R}{\underline R}\rig)^d\sqrt{\frac{d^2}{M}\log\lef(\frac{MR}{d^2}\rig)}.
    \end{align*}
    % Similarly, there exists a multihead ReLU attention layer with number of heads $M\leq\overline R^{\frac{d}{2}}\frac{C(d)}{\epsilon^2}\log\lef(1+C/\epsilon\rig)$, a set of vectors $\{\bfa b_m\}_{m\in[M]}\subset\bb S^{N-1}$ and $\{d_m\}_{m\in[M]}\subset\bb R$ such that 
    % \begin{align*}
    %     \Big|\sum_{m=1}^Md_m\sigma(\bfa b_m^\top\bfa v)-\Vert\bfa v\Vert_2^{1/2}+1\Big|\leq\epsilon.
    % \end{align*}
\end{lemma}
\begin{proof}
    Consider a set $\ca C^d(\overline R) :=\ca B^d_{\infty}(\overline R)\setminus\ca B_{2}^d(\underline R)$, we note that
    \begin{align*}
        \sup_{\bfa v\in\ca C^d(\overline R)}\pta_{v_{j_1},\ldots,v_{j_i}\in[d]}\bl\frac{1}{\Vert\bfa v\Vert_2}\br\leq\frac{C^d}{\Vert\bfa v\Vert_2^d}\leq\frac{C^d}{\underline R^d}.
    \end{align*}
    Therefore, consider the definition \ref{rcsmoothfunc}, we have $C_{\ell}=\lef(\frac{\overline R}{\underline R}\rig)^d C^d$. Then the result is concluded by lemma \ref{Softmaxapprox1/x}.
\end{proof}

\section{Omitted Proofs}

\subsection{Proof of Theorem \ref{thm_main1}}
The proof of the Maximization step follows directly from that of the proof of theorem \ref{thm_main2}. We only change the proof of the Expectation Step for the Transformer layers.
    \begin{center}
    \textbf{1. The Expectation Step.}
\end{center}
For the expectation step, our network is designed by
\begin{align*}
    \bfa V^{(1)}_1&=\begin{bmatrix}
        \bfa 0_{(3d+k+1)\times (D-d)}&\bfa 0\\
        I_d&\bfa 0\\
        \bfa 0&\bfa 0
    \end{bmatrix},\quad\bfa Q^{(1)}_1=\begin{bmatrix}
    \bfa 0_{k\times 2d}&I_k&\bfa 0\\       
        &\bfa 0&
    \end{bmatrix},\quad\bfa K^{(1)}_1=\begin{bmatrix}
        \bfa 0_{d\times(2d+k)}& I_{d}&\bfa 0\\
        &\bfa 0&
    \end{bmatrix},\\
    \bfa V^{(1)}_2&=\begin{bmatrix}
        \bfa 0_{(3d+k+1)\times (D-d)}&\bfa 0\\
        I_d&\bfa 0\\
        \bfa 0&\bfa 0
    \end{bmatrix},\quad\bfa Q^{(1)}_2=\bfa 0,\quad\bfa K^{(1)}_1=\begin{bmatrix}
        \bfa 0_{d\times(2d+k)}& I_{d}&\bfa 0\\
        &\bfa 0&
    \end{bmatrix}.
    % \bfa V_2^{(1)} &= \begin{bmatrix}
    %     \bfa 0_{d\times d}&\bfa 0&\bfa 0\\
    %     \bfa 0_{d\times d}&-I_d &\bfa 0\\
    %     &\bfa 0&
    % \end{bmatrix},\quad\bfa Q_2^{(1)}=\bfa K_2^{(1)}=\begin{bmatrix}
    %     \bfa 0_{d\times d}&\bfa 0&\bfa 0\\
    %     \bfa 0_{d\times d}&-I_d &\bfa 0\\
    %     &\bfa 0&
    % \end{bmatrix}.
\end{align*}
Then we can show that
\begin{align*}
    \bfa V_1^{(1)}\bfa H=\begin{bmatrix}
        &\bfa 0_{d\times N}&\\
        \bfa X_1&\ldots&\bfa X_N\\
        &\bfa 0&
    \end{bmatrix}, \quad\bfa Q_1^{(1)}\bfa H_1=\begin{bmatrix}
        (2\log N)\bfa p_{1,1}^{(0)}&\ldots&(2\log N)\bfa p_{1,N}^{(0)}\\
        &\bfa 0&
    \end{bmatrix},\quad\bfa K_1^{(1)}\bfa H_1=\begin{bmatrix}
        I_d&\bfa 0\\
        \bfa 0&\bfa 0
    \end{bmatrix}.
\end{align*}
Hence we can show that
\begin{align*}
   \softmax\Big( (\bfa Q_1^{(1)}\bfa H_1)^\top(\bfa K_1^{(1)}\bfa H_1)\Big)=\softmax\lef(\begin{bmatrix}
       C(\log N)\bfa p_{1,1}^{(0),\top}&\bfa 0\\
       \vdots&\vdots\\
       C(\log N)\bfa p_{1,N}^{(0),\top}&\bfa 0
   \end{bmatrix}\rig).
\end{align*}
We further note that
\begin{align*}
    \bigg|\frac{\exp(C\log N)}{\sum_{i=1}^N\mbbm 1_{\wh z_i^{(0)}=j}\exp(C\log N)+\mbbm 1_{\wh z_i^{(0)}\neq j}}-\frac{1}{\sum_{i=1}^N\mbbm 1_{\wh z_i^{(0)}=j}}\bigg|\leq N^{-C}.
\end{align*}
Similarly we have
\begin{align*}
    \bigg|\frac{1}{\sum_{i=1}^N\mbbm 1_{\wh z_i^{(0)}=j}\exp(C\log N)+\mbbm 1_{\wh z_i^{(0)}\neq j}}\bigg|\leq N^{-C}.
\end{align*}
We further note that
\begin{align*}
    \softmax\Big( (\bfa Q_1^{(1)}\bfa H_1)^\top(\bfa K_1^{(1)}\bfa H_1)\Big)=\begin{bmatrix}
        \frac{1}{N}&\ldots&\frac{1}{N}\\
        \vdots &\ddots&\vdots\\
        \frac{1}{N}&\ldots&\frac{1}{N}
    \end{bmatrix}.
\end{align*}
Hence one can show that
\begin{align*}
    \lef\Vert\bfa H\sum_{i=1}^2\softmax\lef(\bfa H^\top\bfa Q_i^{(1),\top}\bfa K_i^{(1)}\bfa H\rig)-\begin{bmatrix}
        \wha\mu_1^{(1)}&\ldots&\wha\mu_k^{(1)}&\bfa 0\\
        &\bfa 0
    \end{bmatrix}\rig\Vert_2\leq N^{-1}.
\end{align*}
Therefore, applying $\bfa V_1$ we have
\begin{align*}
   \lef\Vert\sum_{i=1}^2\bfa V_i^{(1)}\bfa H \cdot\softmax\lef(\bfa H^\top\bfa Q_i^{(1),\top}\bfa K_i^{(1)}\bfa H\rig)- \begin{bmatrix}
        \bfa X_1&\bfa X_2&\ldots&\bfa x_{k}&\ldots&\bfa X_N\\
        \wha \mu_{1}^{(0)}&\wha \mu_{2}^{(0)}&\ldots&\wha \mu_{k}^{(0)}&\ldots&\bfa 0\\
        \bfa p^\prime_{1,1}&\bfa p^\prime_{1,2}&\ldots&\bfa p_{1,k}^\prime&\ldots&\bfa p^\prime_{1,N}\\
        \bfa p_{2,1}&\bfa p_{2,2}&\ldots&\bfa p_{2,k}&\ldots&\bfa p_{2,N}\\
        1&1&\ldots&1&\ldots&1\\
        \wha \mu_{1}^{(1)}&\wha \mu_{2}^{(1)}&\ldots&\wha \mu_{k}^{(1)}&\ldots&\bfa 0\\
        &&\bfa 0&&&
    \end{bmatrix}\rig\Vert_2\leq N^{-1}.
\end{align*}
% And we can show that under this construction,
% \begin{align*}
%    \bfa V^{(5)}_1\bfa H_4 (\bfa Q^{(5)}_1\bfa H_4)^\top&=\begin{bmatrix}
%        &&\bfa 0_{d}\\\wha\mu_1^{(1)}&\ldots&\wha\mu_k^{(1)}&\bfa 0\\
%        &&\bfa 0&
%    \end{bmatrix},\quad \bfa K^{(5)}\bfa H_4=\begin{bmatrix}
%         I_d&\bfa 0\\
%         \bfa 0&\bfa 0
%     \end{bmatrix},\\
%     \bfa V^{(5)}_2\bfa H_4(\bfa Q_2^{(5)}\bfa H_4)^\top(\bfa K_2^{(5)}\bfa H_4)&=\begin{bmatrix}
%         &&\bfa 0_{d}&\\
%         \wha\mu_1^{(0)}&\ldots&\wha\mu_k^{(0)}&\bfa 0\\
%         &&\bfa 0&
%     \end{bmatrix}.
% \end{align*}
% And, we can show that
% \begin{align*}
%     \bfa H_{5,1} = \bfa H_4 +\bfa V^{(5)}\bfa H_4(\bfa Q^{(5)}\bfa H_4)^\top(\bfa K^{(5)}\bfa H_4)=\begin{bmatrix}
%         \bfa X_1&\bfa X_2&\ldots&\bfa x_{k}&\ldots&\bfa X_N\\
%         \wha \mu_{1}^{(1)}&\wha \mu_{2}^{(1)}&\ldots&\wha \mu_{k}^{(1)}&\ldots&\bfa 0\\
%         \bfa p^\prime_{1,1}&\bfa p^\prime_{1,2}&\ldots&\bfa p_{1,k}^\prime&\ldots&\bfa p^\prime_{1,N}\\
%         \bfa p_{2,1}&\bfa p_{2,2}&\ldots&\bfa p_{2,k}&\ldots&\bfa p_{2,N}\\
%         1&1&\ldots&1&\ldots&1\\
%         &&\bfa 0&&&
%     \end{bmatrix}.
% \end{align*}
And the Expectation Step is concluded as we updates the centroids from $\{\bfa\mu_i^{(0)}\}_{i\in[k]}$ to $\{\bfa\mu_i^{(1)}\}_{i\in[k]}$. The next step is to update the assignment $\bfa p_{1,i}^\prime$ to $\bfa p_{1,i}^{(1)}$ for $i\in[N]$.
\subsection{Proof of Theorem \ref{thm_main2}}
We first consider the input matrix to be 
\begin{align*}
    \bfa H_1:=\begin{bmatrix}
        \bfa X_1&\bfa X_2&\ldots&\bfa x_n\\
         \wha \mu_{1}^{(0)}&\wha \mu_{2}^{(0)}&\ldots&\bfa 0\\
        \bfa p_{1,1}&\bfa p_{1,2}&\ldots&\bfa p_{1,N}\\
        \bfa p_{2,1}&\bfa p_{2,2}&\ldots&\bfa p_{2,N}\\
        1&1&\ldots&1\\
        \bfa p_{3,1}&\bfa p_{3,2}&\ldots&\bfa p_{3,N}
    \end{bmatrix}\in\bb R^{D\times N}, 
\end{align*}
where $\wh z^{(0)}:[n]\to[k]$ is the assignment function, $\wha\mu_i\in\bb R^{d}$ is the initially estimated centroid for the $i$-th cluster. $\bfa p_{1,i}\in\bb R^k$ satisfies $\bfa p_{1,i,j}=\mbbm 1_{\wh z^{(0)}(i)=j}$ for all $j\in[k]$. And for $\bfa p_{2,i}$ we have $\bfa p_{2,i,j} = \mbbm 1_{j=i}$ for $i\leq d$ and $\bfa p_{2,i,j}=0$ for $i\leq N$ and $j\leq d$. We let $\bfa p_{3,1}=\bfa p_{3,2}=\ldots=\bfa p_{3,N} = \bfa 0\in\bb R^k$.
We note that algorithm \ref{alg:loyld} consists of two iterative steps: (1) The expectation step where we take the averages to get an initial estimate $\wha \mu_{\ell}^{(t)}$. (2) The maximization step where we assign each individual sample their labels. Our following discussions simulate the two steps separately as follows.
\begin{center}
    \textbf{1. The Expectation Step.}
\end{center}
To achieve the first step, we construct our transformer weights as follows:
\begin{align*}
    \bfa V_1^{(1)}=\begin{bmatrix}
        \tda V_{1,1}^{(1)}&\tda V_{1,2}^{(1)}&\tda V_{1,3}^{(1)}
    \end{bmatrix},\quad\bfa Q_1^{(1)}=\begin{bmatrix}
        \bfa 0_{1\times(3d+k)}&1&\bfa 0\\
        \bfa 0&\bfa 0&\bfa 0
    \end{bmatrix},\quad\bfa K_1^{(1)}=\begin{bmatrix}
        \bfa 0_{1\times(3d+k)}&1&\bfa 0\\
        \bfa 0&\bfa 0&\bfa 0
    \end{bmatrix},
\end{align*}
where $\tda V_{1,1}^{(1)}\in\bb R^{2d\times D}=\bfa 0$, $\tda V_{1,2}^{(1)}=\begin{bmatrix}
        \bfa 0_{3d+k}\\
        I_k\\
        \bfa 0
    \end{bmatrix}\in\bb R^{k\times D}$. Then we can show that
\begin{align*}
   \lef(\bfa K_1^{(1)}\bfa H_1\rig)^\top=\lef(\bfa Q_1^{(1)}\bfa H\rig)^\top = \begin{bmatrix}
        1&\bfa 0\\
        \vdots &\bfa 0\\
        1&\bfa 0
    \end{bmatrix}.
\end{align*}
Then we can show that
\begin{align*}
   (\bfa Q_1^{(1)}\bfa H)^\top(\bfa K_1^{(1)}\bfa H) = \begin{bmatrix}
       \bfa v_1&\ldots&\bfa v_1
   \end{bmatrix},\quad \bfa v_{1,i} = 1\quad\forall i\in[N].
\end{align*}
And after the Softmax function we obtain that
\begin{align*}
    \softmax\lef((\bfa Q_1^{(1)}\bfa H)^\top(\bfa K_1^{(1)}\bfa H)\rig)=\begin{bmatrix}
        \frac{1}{D}\bfa v_1&\ldots&\frac{1}{D}\bfa v_1
    \end{bmatrix}.
\end{align*}
Hence, we further obtain that
\begin{align*}
    \bfa V_1^{(1)}\bfa H&\times\softmax((\bfa Q_1^{(1)}\bfa H)^\top(\bfa K_1^{(1)}\bfa H))=\bfa V_1\bfa H\times\begin{bmatrix}
        \frac{1}{D}\bfa v_1 &\ldots&\frac{1}{D}\bfa v_1
    \end{bmatrix}=\bfa V_1\times \begin{bmatrix}
        &\bfa A_0&\\
        \frac{1}{D}\bfa v_k &\ldots&\frac{1}{D}\bfa v_k\\
        &\bfa A_1&
    \end{bmatrix}\\
    &=\begin{bmatrix}
        \tda V_{1,1}^{(1)}&\tda V_{1,2}^{(1)}&\tda V_{1,3}^{(1)}
    \end{bmatrix}\begin{bmatrix}
        &\bfa A_0&\\
        \frac{1}{D}\bfa v_k&\ldots&\frac{1}{D}\bfa v_k\\
        &\bfa A_1&
    \end{bmatrix}=\tda V_{1,2}^{(1)}\begin{bmatrix}
        \frac{1}{D}\bfa v_k &\ldots&\frac{1}{D}\bfa v_k
    \end{bmatrix}\\
    &=\begin{bmatrix}
        \bfa 0_{3d+k}\\
        I_k\\
        \bfa 0
    \end{bmatrix}\begin{bmatrix}
        \frac{1}{D}\bfa v_k&\ldots&\frac{1}{D}\bfa v_k
    \end{bmatrix}=\begin{bmatrix}
        &\bfa 0_{3d+k}&\\
        \frac{1}{D}\bfa v_k&\ldots&\frac{1}{D}\bfa v_k\\
        &\bfa 0&
    \end{bmatrix},
\end{align*}
 where $\bfa A_0\in\bb R^{2d\times N}$ and $
    \bfa v_{k,\ell}=\sum_{i=1}^N\mbbm 1_{\wh z_i^{(0)}=\ell}$.
Then it is checked that
\begin{align*}
   \bfa H_2 =  \bfa H_1+ \bfa V_1^{(1)}\bfa H_1\times\softmax\lef((\bfa Q_1^{(1)}\bfa H_1)^\top(\bfa K_1^{(1)}\bfa H_1)\rig)=\begin{bmatrix}
        \bfa X_1&\bfa X_2&\ldots&\bfa X_N\\
         \wha \mu_{1}^{(0)}&\wha \mu_{2}^{(0)}&\ldots&\bfa 0\\
        \bfa p_{1,1}^{(0)}&\bfa p_{1,2}^{(0)}&\ldots&\bfa p_{1,N}^{(0)}\\
        \bfa p_{2,1}&\bfa p_{2,2}&\ldots&\bfa p_{2,N}\\
        1&1&\ldots&1\\
        \frac{1}{D}\bfa v_k&\frac{1}{D}\bfa v_k&\ldots&\frac{1}{D}\bfa v_k\\
        &&\bfa 0&
    \end{bmatrix}.
\end{align*}
Therefore, we construct the following multi-head layer to remove the off-diagonal elements in $\begin{bmatrix}
    \frac{1}{D}\bfa v_k&\frac{1}{D}\bfa v_k&\ldots&\frac{1}{D}\bfa v_k
\end{bmatrix}$, given by 
\begin{align*}
    \bfa V_{2i}^{(2)} &= -\bfa V_{2i+1}^{(2)}=\frac{N}{N-1}\begin{bmatrix}
        &\bfa 0_{(3d+2k+i)\times D}&\\
        \bfa 0_{1\times(3d+2k+i)}&1&\bfa 0\\
        &\bfa 0&
    \end{bmatrix},\\ 
    {\bfa Q_{2i}^{(2)}}=\bfa K_{2i}^{(2)} &= \begin{bmatrix}
    &\bfa 0_{i-1}&
        \bfa 0_{1\times(2d+k)}&1&\bfa 0\\
        &\bfa 0&
    \end{bmatrix},\quad\bfa K_{2i+1}^{(2)} =\bfa 0,\quad\text{ for }i\in[k].
\end{align*}
Given this formulation, we can show that the mapping $f(\bfa x)= 1$ is $(1,1)$ smooth
\begin{align*}
    \lef(\bfa Q_{2i}^{(2)}\bfa H\rig)^\top\lef(\bfa K_{2i}^{(2)}\bfa H\rig) =\begin{bmatrix}
        \bfa 0_{(i-1)\times 1} &\bfa 0_{1\times (D-1)}\\
        N&\bfa 0_{1\times (D-1)}\\
        0&\bfa 0
    \end{bmatrix}\begin{bmatrix}
        \bfa 0_{1\times (i-1)} &N &\bfa 0\\
        \bfa 0&\bfa 0&\bfa 0
    \end{bmatrix}=\begin{bmatrix}
        &\bfa 0_{(i-1)\times N}&\\
        \bfa 0_{1\times(i-1)}&N^2&\bfa 0\\
        &\bfa 0&
    \end{bmatrix}.
\end{align*}
After the Softmax function, we have
\begin{align*}
   \softmax \lef((\bfa Q_{2i}^{(2)}\bfa H)^\top(\bfa K_{2i}^{(2)}\bfa H)\rig)=\begin{bmatrix}
       \frac{1}{N}&\ldots &\frac{1}{\exp(N^2)+N-1}&\cdots&\frac{1}{N}\\
       \vdots&\ddots&\vdots&\cdots&\vdots\\
       \frac{1}{N}&\cdots &\frac{\exp(N^2)}{\exp(N^2)+N-1}&\cdots&\frac{1}{N}\\
       \vdots&\ddots&\vdots&\ddots&\vdots\\
       \frac{1}{N}&\cdots&\frac{1}{\exp(N^2)+N-1}&\cdots&\frac{1}{N}
   \end{bmatrix}.
\end{align*}
And similarly we can show that $(\bfa Q_{2i+1}^{(2)}\bfa H)^\top(\bfa K_{2i+1}^{(2)}\bfa H)=\bfa 0$, which implies that
\begin{align*}
    \softmax\lef((\bfa Q_{2i+1}^{(2)}\bfa H)^\top(\bfa K_{2i+1}^{(2)}\bfa H)\rig)=\begin{bmatrix}
        \frac{1}{N}&\ldots&\frac{1}{N}\\
        \vdots&\ddots&\vdots\\
        \frac{1}{N}&\ldots&\frac{1}{N}
    \end{bmatrix}.
\end{align*}
We notice that after the softmax, only the $i$-th column remains nonzero, where the value for its $i$-th row is given by
\begin{align*}
    \Big|\frac{\exp(N^2)-1}{\exp(N^2)+D-1}-1\Big|\lesssim D\exp(-N^2).
\end{align*}
Hence, the following holds
\begin{align*}
    \softmax&\lef((\bfa Q_{2i}^{(2)}\bfa H)^\top(\bfa K_{2i}^{(2)}\bfa H)\rig)-\softmax\lef((\bfa Q_{2i+1}^{(2)}\bfa H)^\top(\bfa K_{2i+1}^{(2)}\bfa H)\rig)\\
    &=\begin{bmatrix}
        \bfa 0_{(i-1)\times (i-1)}&\bfa 0_{(i-1)\times 1}&\bfa 0_{(i-1)\times(N-1)}\\
        \bfa 0_{1\times(i-1)}&\frac{N-1}{N}+O\lef( D\exp\lef(-N^2\rig)\rig)&\bfa 0\\
        \bfa 0&\bfa 0&\bfa 0
    \end{bmatrix}\\
    &=\frac{N-1}{N}\begin{bmatrix}
        &\bfa 0_{(3d+2k+i)\times D}&\\
        \bfa 0_{(i-1)}&\bfa v_{k,i} +O(D\exp(-N^2))&\bfa 0\\
        &\bfa 0&
    \end{bmatrix},
\end{align*}
% Hence, we can further show that
% \begin{align*}
%     \bfa V_{i}^{(2)}\bfa H_2\sigma((\bfa Q_i^{(2)}\bfa H_2)^\top(\bfa K_i^{(2)}\bfa H_2)),
% \end{align*}
which immediately implies that
\begin{align*}
    \sum_{i=1}^{2k}\bfa V_i^{(2)}\bfa H_2\softmax\lef((\bfa Q_i^{(2)}\bfa H_2)^\top(\bfa K_i^{(2)}\bfa H_2)\rig)=\begin{bmatrix}
        \bfa 0_{(3d+2k)\times N}&\\
        \diag(\bfa v_k) +O(D\exp(-N^2))&\bfa 0\\
        \bfa 0
    \end{bmatrix}.
\end{align*}
Given the above design, we can show that
\begin{align*}
    \bfa H_{3,1}&=\bfa H_2+\sum_{i=1}^k\bfa V_i^{(2)}\bfa H_2\softmax\lef((\bfa Q_i^{(2)}\bfa H_2)^\top(\bfa K_i^{(2)}\bfa H_2)\rig)\\
    &=\begin{bmatrix}
        \bfa X_1&\bfa X_2&\ldots&\bfa X_N\\
         \wha \mu_{1}^{(0)}&\wha \mu_{2}^{(0)}&\ldots&\bfa 0\\
        \bfa p_{1,1}^{(0)}&\bfa p_{1,2}^{(0)}&\ldots&\bfa p_{1,N}^{(0)}\\
        \bfa p_{2,1}&\bfa p_{2,2}&\ldots&\bfa p_{2,N}\\
        1&1&\ldots&1\\
        \bfa v_k&\bfa v_k&\ldots&\bfa v_k\\
        \diag(\bfa v_k)+O(D\exp(-N^2))&&\bfa 0&\\
        &\bfa 0&&
    \end{bmatrix}.
\end{align*}
Then, we construct the MLP layer to remove the $\bfa v_k$ part, which is designed by
\begin{align*}
    \bfa W_1^{(2)}=I_D,\quad \bfa W_2^{(2)}= \begin{bmatrix}
        &\bfa 0_{(3d+k)\times D}&&\\
        \bfa 0_{k\times (3d+k)}&-I_k&I_k&\bfa 0_{}\\
       \bfa 0 &-I_k&\bfa 0&\bfa 0\\
       &\bfa 0&&
    \end{bmatrix}.
\end{align*}
Given this formulation, we can show that
\begin{align*}
    \bfa H_3 := \bfa H_{3,1}+\bfa W_1^{(2)}\sigma\lef(\bfa W_2^{(2)}\bfa H_{3,1}\rig)=\begin{bmatrix}
        \bfa X_1&\bfa X_2&\ldots&\bfa X_N\\
        \wha \mu_{1}^{(0)}&\wha \mu_{2}^{(0)}&\ldots&\bfa 0\\
        \bfa p_{1,1}^{(0)}&\bfa p_{1,2}^{(0)}&\ldots&\bfa p_{1,N}^{(0)}\\
        \bfa p_{2,1}&\bfa p_{2,2}&\ldots&\bfa p_{2,N}\\
        1&1&\ldots&1\\
        \bfa v_k&\bfa v_k&\ldots&\bfa v_k\\
        \diag(\bfa v_k)+O(D\exp(-N^2))&&\bfa 0&\\
        &\bfa 0&&
    \end{bmatrix}.
\end{align*}

% Then we consider the following attention head which essentially remove the $\bfa v_k$ part in $\bfa H_2$. We note that the rank of $\sigma((\bfa Q_i\bfa H_2)^\top(\bfa K_i\bfa H_2))$ is at most $D$. To form the 
% \begin{align*}
%     \bfa V_{k+1}^{(2)} = \begin{bmatrix}
%         &\bfa 0_{(3d+k)\times D}&\\
%         \bfa 0_{k\times (3d+k)}&-I_k&\bfa 0_{}\\
%         &\bfa 0&
%     \end{bmatrix},\quad \bfa Q_{k+1}^{(2)}=,\quad \bfa K_{k+1}^{(2)}=
% \end{align*}
% Given the above design, we can show that
% \begin{align*}
%     \bfa V_{k+1}^{(2)}\bfa H_2\sigma\lef((\bfa Q_{k+1}^{(2)}\bfa H_2)^\top(\bfa K_i^{(2)}\bfa H_2)\rig)=\begin{bmatrix}
%         &\bfa 0_{(3d+k)\times N}&\\
%         -\bfa v_k&\ldots&-\bfa v_k\\
%         &\bfa 0&
%     \end{bmatrix}.
% \end{align*}
% Therefore, collecting pieces, we can show that
% \begin{align*}
%    \bfa H_3:= \bfa H_2 + \sum_{i=1}^{k+1}\bfa V_i^{(2)}\bfa H_2\sigma\lef((\bfa Q_i^{(2)}\bfa H_2)^\top(\bfa K_i^{(2)}\bfa H_2)\rig)=\begin{bmatrix}
%         \bfa X_1&\bfa X_2&\ldots&\bfa X_N\\
%         \wha \mu_{\wh z^{(0)}(1)}^{(0)}&\wha \mu_{\wh z^{(0)}(2)}^{(0)}&\ldots&\wha \mu_{\wh z^{(0)}(N)}^{(0)}\\
%         \bfa p_{1,1}&\bfa p_{1,2}&\ldots&\bfa p_{1,N}\\
%         \bfa p_{2,1}&\bfa p_{2,2}&\ldots&\bfa p_{2,N}\\
%         1&1&\ldots&1\\
%         \diag(\bfa v_k)&&\bfa 0&\\
%         \bfa p_{4,1}& \bfa p_{4,2}&\ldots &\bfa p_{4,N}
%     \end{bmatrix}.
% \end{align*}
The following layer converts the term $\diag(\bfa v_k)$ to $\diag(\bfa v_k^\prime)$ where $\bfa v_{k,i}^\prime = 1/\bfa v_{k,i}$. The design is given as follows for all $i\in[M]$,
\begin{align*}
    \bfa V_{i}^{(3)} &= \begin{bmatrix}
        &\bfa 0_{(3d+3k+1)\times D}&\\
        \bfa 0_{k\times(2d+2k)}&\diag(c_i)_{k\times k}&\bfa 0\\
        &\bfa 0&
    \end{bmatrix},\quad\bfa Q_{i}^{(3)}=\begin{bmatrix}
        &\bfa 0_{(3d+3k+1)\times D}&\\
        \bfa 0_{k\times(2d+2k)} &I_{k}&\bfa 0\\
        &\bfa 0&
    \end{bmatrix},\\
    \bfa K_{i}^{(3)}&=\begin{bmatrix}
        &\bfa 0_{(3d+3k+1)\times D}&\\
        \bfa 0_{k \times (3d+3k+1)}&\diag(a_i)_{k\times k}&\bfa 0\\
        &\bfa 0&
    \end{bmatrix}\begin{bmatrix}
        &\bfa 0_{(3d+3k+1)\times D}&\\
       \bfa 0_{k\times(3d+2k+1)}&I_{k} &\bfa 0\\
       &\bfa 0&
    \end{bmatrix},
\end{align*}
where we show in lemma \ref{sigmoidapprox} that there exists set of values $\{(c_i,a_i)\}_{i\in[M]}$ such that the following holds for all $\underline R\leq x\leq \overline R$,
\begin{align*}
    \bigg\Vert\sum_{i=1}^Mc_i\sigmoid(a_ix) - \frac{1}{x}\bigg\Vert_2\leq\lef(\frac{C\overline R}{\underline R}\rig)\sqrt{\frac{\log \lef(M \overline R\rig)}{M}}.
\end{align*}
Using the above result, we immediately obtain that
\begin{align*}
    \bfa H_{4,1}:&=\bfa H_3 + \sum_{i=1}^M\bfa V_i^{(3)}\bfa H_3\softmax\lef((\bfa Q_i^{(3)}\bfa H_3)^\top(\bfa K_i^{(3)}\bfa H_3)\rig)\\
    &= \begin{bmatrix}
        \bfa X_1&\bfa X_2&\ldots&\bfa X_N\\
         \wha \mu_{1}^{(0)}&\wha \mu_{2}^{(0)}&\ldots&\bfa 0\\
        \bfa p_{1,1}^{(0)}&\bfa p_{1,2}^{(0)}&\ldots&\bfa p_{1,N}^{(0)}\\
        \bfa p_{2,1}&\bfa p_{2,2}&\ldots&\bfa p_{2,N}\\
        1&1&\ldots&1\\
        \bfa v_k&\bfa v_k&\ldots&\bfa v_k\\
        \diag(\bfa v_k)&&\bfa 0&\\
        \diag(\bfa v_k^\prime)&&\bfa 0&\\
        &\bfa 0&&
    \end{bmatrix}+C\sqrt{\frac{\log(M D)}{M}},
\end{align*}
where $\bfa v_{k,i}^\prime = \bfa v_{k,i}^{-1}$. Then we apply the MLP again with the following design 
\begin{align*}
    \bfa W_1^{(4)}= I_D,\quad\bfa W_2^{(4)}=\begin{bmatrix}
        &\bfa 0_{(3d+k)\times D}&&\\
        \bfa 0_{k\times(3d+k)}&-I_k&I_k&\bfa 0\\
        \bfa 0&-I_k&\bfa 0&\bfa 0\\
        &\bfa 0&&
    \end{bmatrix}.
\end{align*}
The above construction implies that
\begin{align*}
    \bfa H_4 =\bfa W_2^{(3)}\sigma\lef(\bfa W_1^{(3)}\bfa H_3\rig) = \begin{bmatrix}
        \bfa X_1&\bfa X_2&\ldots&\bfa X_N\\
         \wha \mu_{1}^{(0)}&\wha \mu_{2}^{(0)}&\ldots&\bfa 0\\
        \bfa p_{1,1}^{(0)}&\bfa p_{1,2}^{(0)}&\ldots&\bfa p_{1,N}^{(0)}\\
        \bfa p_{2,1}&\bfa p_{2,2}&\ldots&\bfa p_{2,N}\\
        1&1&\ldots&1\\
        \bfa v_k&\bfa v_k&\ldots&\bfa v_k\\
        \diag(\bfa v_k^\prime)&&\bfa 0&\\
        &\bfa 0&&
    \end{bmatrix}.
\end{align*}
We construct the following layer to perform the normalization, given by 
\begin{align*}
    \bfa V_i^{(4)}&= \begin{bmatrix}
    &\bfa 0_{(3d+2k+1)\times D}& \\
    \bfa 0_{k\times(3d+k+1)}&I_{k}&\bfa 0\\
    &\bfa 0&
    \end{bmatrix},\quad \bfa Q_i^{(4)} = \begin{bmatrix}
        &\bfa 0_{(3d+2k+1)\times D}&\\
        \bfa 0_{k\times 3d}&I_{k}&\bfa 0\\
        &\bfa 0&
    \end{bmatrix},\\
    \bfa K_i^{(4)}&=\begin{bmatrix}
        &\bfa 0_{(3d+1)\times D}&\\
        \bfa 0_{d\times (3d+2k+1)}&\bfa A_i&\bfa 0\\
        &\bfa 0&
    \end{bmatrix}\begin{bmatrix}
        &\bfa 0_{(3d+2k+1)\times D}&\\
        \bfa 0_{k\times (3d+1)}&I_k&\bfa 0\\
        &\bfa 0&
    \end{bmatrix}.
\end{align*}
Then, using the approximation bound for the Softmax mapping in lemma \ref{Softmaxapprox1/x}, we can show that there exists $\{(\bfa A_i,c_i)\}_{i\in[M]}$ such that
\begin{align*}
    \sum_{i=1}^Mc_i\softmax\lef((\bfa Q_i^{(4)}\bfa H_4)^\top(\bfa K_i^{(4)}\bfa H_4)\rig) = \begin{bmatrix}
        &\bfa 0_{(3d+2k+1)\times D}&\\
        \bfa p_{1,1}&\ldots&\bfa p_{1,N}\\
        &\bfa 0
    \end{bmatrix}+O\bigg(C^d\sqrt{\frac{d}{M}\log\lef(\frac{M}{d^2}\rig)}\bigg).
\end{align*}
And we also have
\begin{align*}
    \bfa V_2^{(4)}\bfa H_4 = \begin{bmatrix}
        &\bfa 0_{(3d+3k+1)\times D}&\\
        \bfa 0_{k\times(3d+2k+1)}&\diag(\bfa v_k^\prime) +C\sqrt{\frac{\log(MD)}{M}}&\bfa 0\\
        &\bfa 0&
\end{bmatrix},
\end{align*}
which implies that
\begin{align*}
    \bfa H_{4,1} &= \bfa H_3+\sum_{i=1}^M\bfa V_i^{(4)}\bfa H_3\times\softmax\lef((\bfa Q_i^{(4)}\bfa H_3)^\top(\bfa K_i^{(3)}\bfa H_3)\rig)\\
    &=\begin{bmatrix}
        \bfa X_1&\bfa X_2&\ldots&\bfa X_N\\
         \wha \mu_{1}^{(0)}&\wha \mu_{2}^{(0)}&\ldots&\bfa 0\\
        \bfa p_{1,1}^{(0)}&\bfa p_{1,2}^{(0)}&\ldots&\bfa p_{1,N}^{(0)}\\
        \bfa p_{2,1}&\bfa p_{2,2}&\ldots&\bfa p_{2,N}\\
        1&1&\ldots&1\\
        \diag(\bfa v_k^\prime)&&\bfa 0&\\
        \bfa p_{1,1}^\prime&\bfa p_{1,2}^\prime&\ldots&\bfa p_{1,N}^\prime
    \end{bmatrix}+O\bigg(C^d\sqrt{\frac{d}{M}\log\lef(\frac{M}{d^2}\rig)}\bigg),
\end{align*}
where $\bfa p_{1,i}^\prime = \diag(\bfa v_k^\prime)\bfa p_{1,i}$ for all $i\in[N]$. We therefore construct an MLP layer to replace the $\bfa p_{1}$ part using the following design
\begin{align*}
    \bfa W_1^{(4)}=I_D,\qquad\bfa W_2^{(4)} = \begin{bmatrix}
        &&\bfa 0_{2d\times D}&&\\
        \bfa 0_{k\times 2d}&-I_k&\bfa 0_{d\times d}&I_k&\bfa 0\\
        \bfa 0&\bfa 0&\bfa 0&\bfa 0&\bfa 0\\
        \bfa 0_{k\times 2d}&-I_k&\bfa 0&\bfa 0 &\bfa 0\\
        &&\bfa 0_{(d+1)\times D}
    \end{bmatrix},
\end{align*}
which immediately leads to 
\begin{align*}
    \bfa H_4 = \bfa W_1^{(4)}\sigma\lef(\bfa W_2^{(4)}\bfa H_3\rig) +\bfa H_3= \begin{bmatrix}
        \bfa X_1&\bfa X_2&\ldots&\bfa x_{k}&\ldots&\bfa X_N\\
        \wha \mu_{1}^{(0)}&\wha \mu_{2}^{(0)}&\ldots&\wha \mu_{k}^{(0)}&\ldots&\bfa 0\\
        \bfa p^\prime_{1,1}&\bfa p^\prime_{1,2}&\ldots&\bfa p_{1,k}^\prime&\ldots&\bfa p^\prime_{1,N}\\
        \bfa p_{2,1}&\bfa p_{2,2}&\ldots&\bfa p_{2,k}&\ldots&\bfa p_{2,N}\\
        1&1&\ldots&1&\ldots&1\\
        &&\bfa 0&
    \end{bmatrix}+O\bigg(C^d\sqrt{\frac{d}{M}\log\lef(\frac{M}{d^2}\rig)}\bigg).
\end{align*}
Then the next layer is designed as follows
\begin{align*}
    \bfa V^{(5)}_1&=\begin{bmatrix}
        \bfa 0_{d\times (D-d)}&\bfa 0\\
        I_d&\bfa 0\\
        \bfa 0&\bfa 0
    \end{bmatrix},\quad\bfa Q^{(5)}_1=\begin{bmatrix}
    \bfa 0_{k\times 2d}&I_k&\bfa 0\\       
        &\bfa 0&
    \end{bmatrix},\quad\bfa K^{(5)}_1=\begin{bmatrix}
        \bfa 0_{d\times(2d+k)}& I_{d}&\bfa 0\\
        &\bfa 0&
    \end{bmatrix},\\
    \bfa V_2^{(5)} &= \begin{bmatrix}
        \bfa 0_{d\times d}&\bfa 0&\bfa 0\\
        \bfa 0_{d\times d}&-I_d &\bfa 0\\
        &\bfa 0&
    \end{bmatrix},\quad\bfa Q_2^{(5)}=\bfa K_2^{(5)}=\begin{bmatrix}
        \bfa 0_{d\times d}&\bfa 0&\bfa 0\\
        \bfa 0_{d\times d}&-I_d &\bfa 0\\
        &\bfa 0&
    \end{bmatrix}.
\end{align*}
And we can show that under this construction,
\begin{align*}
   \bfa V^{(5)}_1\bfa H_4 (\bfa Q^{(5)}_1\bfa H_4)^\top&=\begin{bmatrix}
       &&\bfa 0_{d}\\\wha\mu_1^{(1)}&\ldots&\wha\mu_k^{(1)}&\bfa 0\\
       &&\bfa 0&
   \end{bmatrix} +O\bigg(C^d\sqrt{\frac{d}{M}\log\lef(\frac{M}{d^2}\rig)}\bigg),\quad \bfa K^{(5)}\bfa H_4=\begin{bmatrix}
        I_d&\bfa 0\\
        \bfa 0&\bfa 0
    \end{bmatrix},\\
    \bfa V^{(5)}_2\bfa H_4(\bfa Q_2^{(5)}\bfa H_4)^\top(\bfa K_2^{(5)}\bfa H_4)&=\begin{bmatrix}
        &&\bfa 0_{d}&\\
        -\wha\mu_1^{(0)}&\ldots&\wha\mu_k^{(0)}&\bfa 0\\
        &&\bfa 0&
    \end{bmatrix}.
\end{align*}
And, we can show that
\begin{align*}
    \bfa H_{5,1} = \bfa H_4 +\bfa V^{(5)}\bfa H_4(\bfa Q^{(5)}\bfa H_4)^\top(\bfa K^{(5)}\bfa H_4)=\begin{bmatrix}
        \bfa X_1&\bfa X_2&\ldots&\bfa x_{k}&\ldots&\bfa X_N\\
        \wha \mu_{1}^{(1)}&\wha \mu_{2}^{(1)}&\ldots&\wha \mu_{k}^{(1)}&\ldots&\bfa 0\\
        \bfa p^\prime_{1,1}&\bfa p^\prime_{1,2}&\ldots&\bfa p_{1,k}^\prime&\ldots&\bfa p^\prime_{1,N}\\
        \bfa p_{2,1}&\bfa p_{2,2}&\ldots&\bfa p_{2,k}&\ldots&\bfa p_{2,N}\\
        1&1&\ldots&1&\ldots&1\\
        &&\bfa 0&&&
    \end{bmatrix}.
\end{align*}
And the Expectation Step is concluded as we update the centroids from $\{\bfa\mu_i^{(0)}\}_{i\in[k]}$ to $\{\bfa\mu_i^{(1)}\}_{i\in[k]}$. The next step is to update the assignment $\bfa p_{1,i}^\prime$ to $\bfa p_{1,i}^{(1)}$ for $i\in[N]$.
\begin{center}
    \textbf{2. The Maximization Step.}
\end{center}
The maximization step involves two sub-networks:
\textbf{Step 1:} Copy the $\bfa x$ below the $1$s, yielding $$\begin{bmatrix}
        \bfa X_1&\bfa X_2&\ldots&\bfa x_{k}&\ldots&\bfa X_N\\
        \wha \mu_{1}^{(1)}&\wha \mu_{2}^{(1)}&\ldots&\wha \mu_{k}^{(1)}&\ldots&\bfa 0\\
        \bfa p^\prime_{1,1}&\bfa p^\prime_{1,2}&\ldots&\bfa p_{1,k}^\prime&\ldots&\bfa p^\prime_{1,N}\\
        \bfa p_{2,1}&\bfa p_{2,2}&\ldots&\bfa p_{2,k}&\ldots&\bfa p_{2,N}\\
        1&1&\ldots&1&\ldots&1\\
        \bfa x_{1,1}&\bfa x_{2,1}&\ldots&\bfa x_{k,1}&\ldots&\bfa x_{N,1}\\
        &&\vdots\\
        \bfa x_{1,k}&\bfa x_{2,k}&\ldots&\bfa x_{k,k}&\ldots&\bfa x_{N,k}\\
        &&\bfa 0
    \end{bmatrix};$$
\textbf{Step 2:} Move $\{\wha\mu_i^{(1)}\}_{i\in[M]}$ to $\{\bfa x_{j,i}\}_{j\in[N]}$, yielding
$$\begin{bmatrix}
        \bfa X_1&\bfa X_2&\ldots&\bfa x_{k}&\ldots&\bfa X_N\\
        \wha \mu_{1}^{(1)}&\wha \mu_{2}^{(1)}&\ldots&\wha \mu_{k}^{(1)}&\ldots&\bfa 0\\
        \bfa p^\prime_{1,1}&\bfa p^\prime_{1,2}&\ldots&\bfa p_{1,k}^\prime&\ldots&\bfa p^\prime_{1,N}\\
        \bfa p_{2,1}&\bfa p_{2,2}&\ldots&\bfa p_{2,k}&\ldots&\bfa p_{2,N}\\
        1&1&\ldots&1&\ldots&1\\
        \bfa x_{1,1}-\wha\mu_1^{(1)}&\bfa x_{2,1}-\wha \mu_1^{(1)}&\ldots&\bfa x_{k,1}-\wha\mu_1^{(1)}&\ldots&\bfa x_{N,1}-\wha\mu_1^{(1)}\\
        &&\vdots\\
        \bfa x_{1,k}-\wha\mu_k^{(1)}&\bfa x_{2,k}-\wha\mu_k^{(1)}&\ldots&\bfa x_{k,k}-\wha\mu_k^{(1)}&\ldots&\bfa x_{N,k}-\wha\mu_k^{(1)}\\
        &&\bfa 0
    \end{bmatrix};$$ \textbf{Step 3:} One-by-one, compute the norm and obtain the following matrix
    $$\begin{bmatrix}
        \bfa X_1&\bfa X_2&\ldots&\bfa x_{k}&\ldots&\bfa X_N\\
        \wha \mu_{1}^{(1)}&\wha \mu_{2}^{(1)}&\ldots&\wha \mu_{k}^{(1)}&\ldots&\bfa 0\\
        &&\bfa 0_{k\times N}
        \\
        \bfa p_{2,1}&\bfa p_{2,2}&\ldots&\bfa p_{2,k}&\ldots&\bfa p_{2,N}\\
        1&1&\ldots&1&\ldots&1\\
         \Vert\bfa x_{1,1}-\wha\mu_1^{(1)}\Vert_2&\Vert\bfa x_{2,1}-\wha \mu_1^{(1)}\Vert_2&\ldots&\Vert\bfa x_{k,1}-\wha\mu_1^{(1)}\Vert_2&\ldots&\Vert\bfa x_{N,1}-\wha\mu_1^{(1)}\Vert_2\\
        &&\vdots\\
        \Vert\bfa x_{1,k}-\wha\mu_k^{(1)}\Vert_2&\Vert\bfa x_{2,k}-\wha\mu_k^{(1)}\Vert_2&\ldots&\Vert\bfa x_{k,k}-\wha\mu_k^{(1)}\Vert_2&\ldots&\Vert\bfa x_{N,k}-\wha\mu_k^{(1)}\Vert_2\\
        &&\bfa 0
    \end{bmatrix};$$ \textbf{Step 4:} And finally, we apply the Softmax and recover approximates to $\{\tda p_{1,i}^{(2)}\}_{i\in[N]}$. Then we move it back to the original places belonging to $\{\bfa p_{1,i}^\prime\}_{i\in[k]}$. Then we give the following construction for each step:
    \begin{center}
        \textbf{Step 1.}
    \end{center}
    In step 1, we construct the following parameters
    \begin{align*}
        \bfa W^{(5)}_1=\begin{bmatrix}
            \bfa 0_{(3d+k+1)\times N}\\
            I_d&\bfa 0\\
            \bfa 0
        \end{bmatrix},\quad\bfa W_2^{(5)} = I_D.
    \end{align*}
    And the Attention layer of the $6$-th layer makes an identity mapping.
    Then we consider 
    \begin{align*}
        \bfa W^{(6)}_1=\begin{bmatrix}
            \bfa 0_{(3d+k+1)\times N}\\
            I_d&\bfa 0\\
            \bfa 0
        \end{bmatrix},\quad \bfa W_2^{(5)}=-I_D.
    \end{align*}
    Similarly we define $\lef\{\bfa W_1^{(i)},\bfa W_2^{(i)}\rig\}_{i\in[7:6+2(k-1)]}$, and the task is achieved. Hence we show that
    \begin{align*}
        \bfa H_{4+2k}=\begin{bmatrix}
        \bfa X_1&\bfa X_2&\ldots&\bfa x_{k}&\ldots&\bfa X_N\\
        \wha \mu_{1}^{(1)}&\wha \mu_{2}^{(1)}&\ldots&\wha \mu_{k}^{(1)}&\ldots&\bfa 0\\
        \bfa p^\prime_{1,1}&\bfa p^\prime_{1,2}&\ldots&\bfa p_{1,k}^\prime&\ldots&\bfa p^\prime_{1,N}\\
        \bfa p_{2,1}&\bfa p_{2,2}&\ldots&\bfa p_{2,k}&\ldots&\bfa p_{2,N}\\
        1&1&\ldots&1&\ldots&1\\
        \bfa x_{1,1}&\bfa x_{2,1}&\ldots&\bfa x_{k,1}&\ldots&\bfa x_{N,1}\\
        &&\vdots\\
        \bfa x_{1,k}&\bfa x_{2,k}&\ldots&\bfa x_{k,k}&\ldots&\bfa x_{N,k}\\
        &&\bfa 0
    \end{bmatrix}.
    \end{align*}
    \begin{center}
        \textbf{Step 2.}
    \end{center}
    To achieve step 2, we note that the following holds
    \begin{align*}
    \tda V^{(j)}\bfa H_{4+2k}\tda G^{(j)}= \begin{bmatrix}
        &&\bfa 0_{k+3d+j}\\
        -\wha\mu_j^{(1)}&-\wha \mu_j^{(1)}&\ldots&-\wha\mu_j^{(1)}\\
        &&\bfa 0\\
    \end{bmatrix}=\Delta\bfa H_{0},
    \end{align*}
    where 
    \begin{align*}
        \tda V^{(j)} =\begin{bmatrix}
            &\bfa 0_{(3d+k+j)\times D}&\\
            0_{d\times 1}&I_d&\bfa 0\\
            &\bfa 0&
        \end{bmatrix},\quad\tda G=\begin{bmatrix}
            1&\ldots&1\\
            &\bfa 0&
        \end{bmatrix}.
    \end{align*}
    Hence, using lemma \ref{softmaxapprox} to show that there exists a set of parameter $\lef\{\bfa V_i^{(5+2k)},\bfa Q_i^{(5+2k)},\bfa K_i^{(5+2k)}\rig\}_{i\in[M}$ given as follows for $i\in[M]$,
    \begin{align*}
        \bfa V_i^{(5+2k)}=c_i\tda V,\quad \bfa Q_i^{(5+2k)} = \begin{bmatrix}
            \bfa 0_{(2d+k)\times d}&I_d&\bfa 0\\
            &\bfa 0&
        \end{bmatrix},\quad \bfa K_i^{(5k+2)} =\begin{bmatrix}
            \bfa A_i&\bfa 0\\
            \bfa 0&\bfa 0
        \end{bmatrix}\times\begin{bmatrix}
            \bfa 0_{1\times (3d+k+1)}&1&\bfa 0\\
            &\bfa 0&
        \end{bmatrix}.
    \end{align*}
    And one can show that
    \begin{align*}
        \bigg\Vert\Delta \bfa H_1- \sum_{i=1}^M\bfa V_i\bfa H^{(4+2k)}\softmax\lef(\bfa H_i^{(2k+4),\top}\bfa Q_i^\top\bfa K_i^{(2k+4)}\bfa H^{(4+2k)}_i\rig)\bigg\Vert_{2}\leq \Vert\wha\mu_1^{(1)}\Vert_2\sqrt{\frac{d}{M}\log(M)}.
    \end{align*}
    Hence, we design accordingly the next $(k-1)M$ heads $\lef\{\bfa V_i^{(2k+4)},\bfa Q_i^{(2k+4)},\bfa K_i^{(2k+4)}\rig\}_{j\in[2:k]}$ similarly. We also let all the FC layer preserve their identity maps. The above construction implies that
    \begin{align*}
        \Big\Vert\sum_{i=1}^k\Delta \bfa H_i-\sum_{i=1}^{kM}\bfa V_i\bfa H^{(4+2k)}\softmax\lef(\bfa H_i^{(2k+4),\top}\bfa Q_i^\top\bfa K_i^{(2k+4)}\bfa H_i\rig)\Big\Vert_2\leq \sqrt k\sup_{j\in[k]}\lef\Vert\wha\mu_j^{(1)}\rig\Vert_2\sqrt{\frac{d\log (M)}{M}},
    \end{align*}
    which alternatively implies that
    \begin{align*}
       \lef\Vert \bfa H_{4+3k}-\begin{bmatrix}
        \bfa X_1&\bfa X_2&\ldots&\bfa x_{k}&\ldots&\bfa X_N\\
        \wha \mu_{1}^{(1)}&\wha \mu_{2}^{(1)}&\ldots&\wha \mu_{k}^{(1)}&\ldots&\bfa 0\\
        \bfa p^\prime_{1,1}&\bfa p^\prime_{1,2}&\ldots&\bfa p_{1,k}^\prime&\ldots&\bfa p^\prime_{1,N}\\
        \bfa p_{2,1}&\bfa p_{2,2}&\ldots&\bfa p_{2,k}&\ldots&\bfa p_{2,N}\\
        1&1&\ldots&1&\ldots&1\\
        \bfa x_{1,1}-\wha\mu_1^{(1)}&\bfa x_{2,1}-\wha\mu_1^{(1)}&\ldots&\bfa x_{k,1}-\wha\mu_1^{(1)}&\ldots&\bfa x_{N,1}-\wha\mu_1^{(1)}\\
        &&\vdots\\
        \bfa x_{1,k}-\wha\mu_k^{(1)}&\bfa x_{2,k}-\wha\mu_k^{(1)}&\ldots&\bfa x_{k,k}-\wha\mu_k^{(1)}&\ldots&\bfa x_{N,k}-\wha\mu_k^{(1)}\\
        &&\bfa 0
    \end{bmatrix} \rig\Vert_{2}\lesssim \sqrt k\sup_{j\in[k]}\lef\Vert\wha\mu_j^{(1)}\rig\Vert_2\sqrt{\frac{d\log(M)}{M}}.
    \end{align*}
    \begin{center}
        \textbf{Step 3.}
    \end{center}
    To achieve step 3, we first divide the heads into $k$-blocks where the $j$-th block achieves the task of approximating the norm function $\Vert\bfa x_{1,j}-\wha\mu_j^{(1)}\Vert_2$ taking $\bfa x_{1,j}-\wha\mu_j^{(1)}$ as input. We design the parameters for the first block as follows, as an example
    \begin{align*}
        \bfa V_i^{(5+3k)}&= \begin{bmatrix}
            &\bfa 0_{(k+3)d+k+1}&\\
            \bfa 0_{d\times((k+3)d+k+1)}&c_iI_d&\bfa 0\\
            &\bfa 0&
        \end{bmatrix},\quad \bfa Q_i^{(5+3k)}=\begin{bmatrix}
            &\bfa 0_{(2d+k)\times N}&\\
            \bfa 0_{2d+k}& I_d&\bfa 0\\
            &\bfa 0&
            \end{bmatrix},\\
        \bfa K_i^{(5+3k)}&=\begin{bmatrix}
            &\bfa 0_{(3d+k+1)\times D}&\\
            \bfa 0_{d\times(3d+k+1)}&\bfa A_i&\bfa 0\\
            &\bfa 0&
        \end{bmatrix}\cdot\begin{bmatrix}
            &\bfa 0_{(3d+k+1)\times D}&\\
            \bfa 0_{d\times(3d+k+1)}&I_d&\bfa 0\\
            &\bfa 0&
        \end{bmatrix}.
    \end{align*}
      $\{c_i,\bfa A_i\}_{i\in[M]}$ satifies $\sup_{i\in[M]}\Vert\bfa A\Vert_2\leq C$, whose existence is guaranteed by lemma \ref{softmaxapprox}. Moreover, it is not hard to show that one can utilize 2 FC layers to remove the vector part and the $\{\bfa p_{1,i}^\prime\}_{i\in[N]}$ Under this design, our final output satisfies
      \begin{align*}
       &\lef\Vert \bfa H_{6+3k}-\begin{bmatrix}
        \bfa X_1&\bfa X_2&\ldots&\bfa x_{k}&\ldots&\bfa X_N\\
        \wha \mu_{1}^{(1)}&\wha \mu_{2}^{(1)}&\ldots&\wha \mu_{k}^{(1)}&\ldots&\bfa 0\\
        &&\bfa 0_{k\times N}
        \\
        \bfa p_{2,1}&\bfa p_{2,2}&\ldots&\bfa p_{2,k}&\ldots&\bfa p_{2,N}\\
        1&1&\ldots&1&\ldots&1\\
         \Vert\bfa x_{1,1}-\wha\mu_1^{(1)}\Vert_2&\Vert\bfa x_{2,1}-\wha \mu_1^{(1)}\Vert_2&\ldots&\Vert\bfa x_{k,1}-\wha\mu_1^{(1)}\Vert_2&\ldots&\Vert\bfa x_{N,1}-\wha\mu_1^{(1)}\Vert_2\\
        &&\vdots\\
        \Vert\bfa x_{1,k}-\wha\mu_k^{(1)}\Vert_2&\Vert\bfa x_{2,k}-\wha\mu_k^{(1)}\Vert_2&\ldots&\Vert\bfa x_{k,k}-\wha\mu_k^{(1)}\Vert_2&\ldots&\Vert\bfa x_{N,k}-\wha\mu_k^{(1)}\Vert_2\\
        &&\bfa 0
    \end{bmatrix} \rig\Vert_{2}\\
    &\lesssim \sqrt k\sup_{j\in[k]}\lef\Vert\wha\mu_j^{(1)}\rig\Vert_2\sqrt{\frac{\log(M)}{M}}.
    \end{align*}
    \begin{center}
        \textbf{Step 4.}
    \end{center}
    In step 4, we utilize the property of the Softmax function to approximate the hard max
    and replace the $\bfa p_{1,1}^\prime$ with our new estimate $\bfa p_{1,k}^{(1)}$. We construct our layer weights by
    \begin{align*}
        \bfa V^{(7+3k)}&= \begin{bmatrix}
            &\bfa 0_{(2d)\times N}&\\
            \bfa 0_{k\times (2d+k)}&I_k&\bfa 0\\
            &\bfa 0&
        \end{bmatrix},\quad\bfa Q^{(7+3k)} = \begin{bmatrix}
            &\bfa 0_{(2d+k)}&
            \bfa 0_{(2d+k)\times k}&I_k&\bfa 0\\
             &\bfa 0&
        \end{bmatrix},\\
        \bfa K^{(7+3k)}&=C\log N\begin{bmatrix}
            &\bfa 0_{(3d+k+1)\times N}&\\
            \bfa 0_{k\times(2d)}&I_k&\bfa 0\\
            &\bfa 0&
        \end{bmatrix}.
    \end{align*}
    Note that the result given by \ref{approxhardmax} implies that
    \begin{align*}
        \lef\Vert\softmax\lef(\begin{bmatrix}
         \Vert\bfa x_{1,1}-\wha\mu_1^{(1)}\Vert_2&\ldots&\Vert\bfa x_{N,1}-\wha\mu_1^{(1)}\Vert_2\\
        &\ddots&\\
        \Vert\bfa x_{1,k}-\wha\mu_k^{(1)}\Vert_2&\ldots&\Vert\bfa x_{N,k}-\wha\mu_k^{(1)}\Vert_2
        \end{bmatrix}\rig)-\begin{bmatrix}
            \bfa p_{1,1}^{(1)}&\bfa p_{1,2}^{(1)}&\ldots&\bfa p_{1,N}^{(1)}
        \end{bmatrix}\rig\Vert_2\lesssim N\exp(-C\log N) = dN^{-C}.
    \end{align*}
     Under this construction and let the FC layer retain the identity of the first $3d+k+1$ columns and remove the rest, our approximation results in lemma \ref{approxhardmax} implies that
    \begin{align*}
        \lef\Vert\bfa H_{7+3k}- \begin{bmatrix}
        \bfa X_1&\bfa X_2&\ldots&\bfa x_{k}&\ldots&\bfa X_N\\
        \wha \mu_{1}^{(1)}&\wha \mu_{2}^{(1)}&\ldots&\wha \mu_{k}^{(1)}&\ldots&\bfa 0\\
        \bfa p_{1,1}^{(1)}&\bfa p_{1,2}^{(1)}&\ldots&\bfa p_{1,k}^{(1)}&\ldots&\bfa p_{1,N}^{(1)}
        \\
        \bfa p_{2,1}&\bfa p_{2,2}&\ldots&\bfa p_{2,k}&\ldots&\bfa p_{2,N}\\
        1&1&\ldots&1&\ldots&1\\
        &&\bfa 0
    \end{bmatrix}\rig\Vert_2\lesssim dN^{-C}+\sqrt k\sup\Vert\wha\mu_j^{(1)}\Vert_2\sqrt{\frac{\log M}{M}}+C^d\sqrt{\frac{d^2\log M}{M}}.
    \end{align*}
    Hence, we construct in total of $\tau$ sets of subnetwork, and use the final layer to extract the set of output $\{\bfa p_{1,i}^{(\tau)}\}_{i\in[N]}$. By subadditivity of the $L_2$ norm we finalize our results by
    \begin{align*}
        \Big\Vert\tfp(\bfa H)&-\begin{bmatrix}
            \bfa p_{1,1}^{(1)}&\bfa p_{1,2}^{(1)}&\ldots&\bfa p_{1,N}^{(1)}
        \end{bmatrix}\Big\Vert_2\lesssim \tau \Big(dN^{-C}+\sqrt k\sup\Vert\wha\mu_j^{(1)}\Vert_2\sqrt{\frac{\log M}{M}}+C^d\sqrt{\frac{d^2\log M}{M}}\Big).
    \end{align*}
% \begin{lemma}[Identity Mappings of the Softmax Function]
%     There exists a set of weights $((\bfa v_1, a_1),\ldots,(\bfa v_M,a_M))\in[\bb R^{d+1}\times[-B,B]]^M$. The mapping $f_i(\bfa x):=x_i$ can be approximated by a sum of the Softmax functions with the number of heads $M<$ such that
%     \begin{align*}
%         \sup_{0\leq\alpha\leq m}\Big|f(\bfa x)-\frac{1}{M}\Big|\leq\epsilon
%     \end{align*}
% \end{lemma}
% \begin{proof}
%     By lemma \ref{lm:a1} we can show that for $M\gtrsim C^2(d)B_f^2/\epsilon^2$ there exists $((\bfa v_1, a_1),\ldots,(\bfa v_M,a_M))\in[\bb R^{d+1}\times[-B,B]]^M$ such that the function $f(\bfa x)=x_i$ can be approximated with
%     \begin{align*}
%         \sup_{\bfa x\in U}\Big|f(\bfa x)-\frac{1}{M}\sum_{i=1}^Ma_i\sigmoid\lef([\bfa x^\top, 1]\bfa v_i\rig)\Big|\leq\epsilon.
%     \end{align*}
% \end{proof}
\begin{lemma}[Approximating the Hardmax Function by the Softmax Function]\label{approxhardmax} Consider a vector $\bfa x\in\bb R^d$. Let $x^*=\max_{i}\{x_i\}_{i\in[d]}$. Define $x^*=\max_{i\in[d]}x_i$, $\ca N_2=\{i:x_i=x^*\}$, $\Delta = x^*-\max_{i\in\ca N_2}x_i$. Define the Hardmax function as $\text{Hardmax}(\bfa x)=\begin{bmatrix}
    \frac{\mbbm 1_{x_1=\max_{i\in[d]}\{x_i\}}}{\sum_{i=1}^d\mbbm 1_{x_i=\max_{i\in[d]}}\{x_i\}}&\ldots&\frac{\mbbm 1_{x_d=\max_{i\in[d]}\{x_i\}}}{\sum_{i=1}^d\mbbm 1_{x_i=\max_{i\in[d]}}\{x_i\}}
\end{bmatrix}$, we subsequently show that 
\begin{align*}
    \Big|\softmax(\beta\bfa v)-\text{Hardmax}(\bfa v)\Big|\leq\bigg((d-|\ca N_2|)+\frac{(d-|\ca N_2|)^2}{|\ca N_2|^3}\bigg)^{\frac{1}{2}}\exp(-\beta\Delta).
\end{align*}
\end{lemma}
\begin{proof}
 We can show that the difference is given by
    \begin{align*}
        \Big\Vert\softmax(&\beta \bfa v)-\text{Hardmax}(\bfa v)\Big\Vert_{2} =\bigg(\sum_{i=1}^d\bigg(\frac{\mbbm 1_{x_i=\max_{i\in[d]}\{x_i\}}}{\sum_{i=1}^d\mbbm 1_{x_i=\max_{i\in[d]}\{x_i\}}}-\frac{\exp(\beta x_i)}{\sum_{i=1}^d\exp(\beta x_i)}\bigg)^2\bigg)^{\frac{1}{2}} \\
        &=\bigg(\sum_{i=1}^d\bigg(\frac{\mbbm 1_{x_i=\max_{i\in[d]}\{x_i\} }}{\sum_{i=1}^d\mbbm 1_{x_i=\max_{i\in[d]}\{x_i\} }}-\frac{\exp(\beta(x_i-x^*))}{\sum_{i=1}^d\exp(\beta(x_i-x^*)) }\bigg)\bigg)^{\frac{1}{2}}\\
        &\leq\bigg(\sum_{i=1}^d\mbbm 1_{x_i\neq x^*}\exp\lef(2\beta(x_i-x^*)\rig)+\mbbm 1_{x_i=x^*}\bl\frac{\exp\lef(\beta(x_i-x^*)\rig)}{\sum_{i=1}^d\exp\lef(\beta(x_i-x^*)
        \rig)}-\frac{1}{|\{i:x_i=x^*\}|}\br^2\bigg)^{\frac{1}{2}}\\
        &\leq \Big(|\ca N_1|\exp(-2\beta\Delta)+ |\ca N_2|\Big(\frac{1}{|\ca N_2|+|\ca N_1|\exp(-\beta \Delta)}-\frac{1}{|\ca N_2|}\Big)^2\Big)^{\frac{1}{
        2}}\\
        &\leq \bigg((d-|\ca N_2|)\exp(-2\beta\Delta) +\frac{(d-|\ca N_2|)^2\exp(-2\beta\Delta)}{|\ca N_2|\lef(|\ca N_2|+(d-|\ca N_2|)\exp(-\beta\Delta)\rig)^2}\bigg)^{\frac{1}{2}}\\
        &=\bigg((d-|\ca N_2|)+\frac{\lef(d-|\ca N_2|\rig)^2}{|\ca N_2|\lef(|\ca N_2|+(d-|\ca N_2|)\exp(-\beta\Delta)\rig)^2}\bigg)^{\frac{1}{2}}\exp\lef(-\beta\Delta\rig)\\
        &\leq \bigg((d-|\ca N_2|)+\frac{(d-|\ca N_2|)^2}{|\ca N_2|^3}\bigg)^{\frac{1}{2}}\exp(-\beta\Delta).
    \end{align*}
\end{proof}

\subsection{Proof of Proposition \ref{genbd}}
    Before we start the proof, we first consider the event of $\ca E=\lef\{\lef\Vert\bfa X_i^{(j)}-\bfa\mu_{\bfa z_i}^{(j)}\rig\Vert_2\leq \sigma\sqrt{d\log(nN/\delta)},\forall i\in[N],j\in[n]\rig\}$, it is not hard to check that by sub-Gaussian tail bound, $\bb P(E)\geq 1-\delta$.
    Then, it is not hard to check that under $E$, $\Vert \bfa H\Vert_2\lesssim \sigma\sqrt{d\log(nN/\delta)}\times N$.    
    Then we obtain generalization bound under the event $\ca E$ using the proof machine created by \citet{bai2024transformers} J.2 \citet{he2025learning} Proposition 1, where we note that the multi-layered Transformer satisfies the following conditions
    \begin{enumerate}
        \item The metric entropy of the operator norm ball satisfies $\log\lef(\delta, \ca B_{\vertiii \cdot}, \vertiii\cdot\rig)\leq C B_LB_MD^2\log(1+(B_{\bfa\theta}+B_X+k)/\delta)$.
        \item $L(TF_{\bfa\theta}(\bfa H), \bfa P_1(\bfa z))\leq 2$ and is $2$ sub-Gaussian.
        \item The Lipschitz condition of the Transformer satisfies for $\bfa\theta_1,\bfa\theta_2\in\Theta_{B_M,B_L}(B_{\bfa\theta})$,
        \begin{align*}
         L(TF_{\bfa\theta_1}(\bfa H), \bfa P_1(\bfa z))- L(TF_{\bfa\theta_2}(\bfa H), \bfa P_1(\bfa z))\leq CLB_{\bfa\theta}^{4L}B_X^{3L}\vertiii{\bfa\theta_1-\bfa\theta_2}.
        \end{align*}
    \end{enumerate}
    where the last condition is obtained through noticing that the Softmax function is Lipschitz with constant $C$. And for bounded input $\bfa H$ the Lipschitz constant of the non-activated Attention layer is proportional to $B_{\bfa\theta}^3$. Then, using proposition A.4 in \citet{bai2024transformers}, one can show by union bound that with probability at least $1-\delta$,
    \begin{align*}
        L\lef(A_{\wha\theta}(\bfa H), \bfa P_1(\bfa z)\rig)& \leq \inf_{\bfa\theta\in\Theta_{B_M,B_L}(B_{\bfa\theta})}\bb E[L\lef(A_{\wha\theta}(\bfa H), \bfa P_1(\bfa z)\rig)|\ca E]\\
        &+C\sqrt{\frac{D^2B_LB_M\log(NB_{\bfa\theta}B_{M}D\sigma\cdot\sup_{i\in[N],j\in[n]}\Vert\bfa\mu_i^{(j)}\Vert_2)+\log(2/\delta)}{n}},
    \end{align*}
    where $A\in\{TF,TF^+\}$.

\subsection{Proof of Theorem \ref{ultimate}}
    We prove the above theorem through upper bounding the term $\inf_{\bfa\theta\in\Theta_{B_M,B_L(B_{\bfa\theta})}}\bb E[L(A_{\wha\theta}(\bfa H),\bfa P_1(\bfa z))|\ca A\cap\ca E]$ where $\ca A$ appears to be the event in Corollary 3 in \citet{lu2016statistical}. Note that $\bb P(\ca A)\geq 1-5n^{-1}-2\exp(-\Delta/\sigma)$. Since the proof for the Transformer and the Transformer+ shares similar idea, we only prove the case with the Transformer model where $\tda\theta$ is the construction. Denote the estimate given by Lloyd's algorithm as $
    \wha P(\bfa z):=\begin{bmatrix}
        \bfa p_{1,1}^{(\tau)}&\ldots&\bfa p_{1,N}^{(\tau)}
    \end{bmatrix}$ Then we can see that
    \begin{align*}
\inf_{\bfa\theta\in\Theta_{B_M,B_L(B_{\bfa\theta})}}\bb E&[L(TF_{\wha\theta}(\bfa H),\bfa P_1(\bfa z))|\ca A\cap\ca E]\leq \bb E\lef[L(TF_{\tda\theta}(\bfa H),\bfa P_1(\bfa z))|\ca A\cap\ca E\rig]\\
&\leq \bb E\lef[L(\wha P_1(\bfa z), \bfa P_1(\bfa z))|\ca A\cap\ca E\rig]+\frac{1}{N}\Vert\wha P_1-TF_{\tda\theta}(\bfa H)\Vert_1.
    \end{align*}
    Note that the first term can be controlled by \citet{yu2015useful} Corollary 3.1 as follows
    \begin{align*}
        \bb E\lef[L(\wha P_1(\bfa z),\bfa P_1(\bfa z))\Big|\ca E\rig]&\leq 2\bb E\Big[L(\wha P_1(\bfa z),\bfa P_1(\bfa z))|\ca A\cap \ca E\Big]\bb P(\ca A)+2\bb P(\ca A)\\
        &\leq \exp\lef(-(1+o(1))\frac{\Delta^2}{8\sigma^2}\rig) +\frac{10}{n}+2\exp\lef(-\frac{\Delta}{\sigma}\rig).
    \end{align*}
    For the second term, we note that
    \begin{align*}
        \frac{1}{N}\lef\Vert\wha P_1(\bfa z)-TF_{\tda\theta}(\bfa H)\rig\Vert_1\leq\frac{1}{\sqrt N}\lef\Vert\wha P_1(\bfa z)-TF_{\tda\theta}(\bfa H)\rig\Vert_2.
    \end{align*}
Then we can show that, with probability at least $1-\delta - 5n^{-1} - 2\exp(-\Delta/\sigma)$, given $\tau\geq 4\log n+1$, $B_L= k\tau+C\tau$, $B_M\geq M$, $D=Cdk$,
\begin{align*}
    L\lef(TF_{\wha\theta}(\bfa H),\bfa P_1(\bfa z)\rig) &\lesssim \exp\lef(-(1+o(1))\frac{\Delta^2}{8\sigma^2}\rig)\\
&+\sqrt{\frac{D^2B_LB_M\log(NB_{\bfa\theta}B_{M}D\sigma\cdot\sup_{i\in[N],j\in[n]}\Vert\bfa\mu_i^{(j)}\Vert_2)+\log(2/\delta)}{n}}\\
&+\tau \sqrt N C^d\Big(\sqrt k\sup_{j\in[k],\ell\in[\tau]}\Vert\wha\mu_j^{(1)}\Vert_2\sqrt{\frac{\log M}{M}}+N^{-1}\Big).
\end{align*}
We further let $M=n^{\frac{1}{4}}, L\asymp k\log n$ and obtain that with probability at least $1-\delta - 5n^{-1} - 2\exp(-\Delta/\sigma)$.
\begin{align*}
    L(TF_{\wha\theta}(\bfa H),\bfa P_1(\bfa z))\lesssim \exp\lef(-(1+o(1))\frac{\Delta^2}{8\sigma^2}\rig)+\sqrt kn^{-1/4}C^d\sqrt{Polylog(n)+\log(2/\delta)}+N^{-3/2}.
\end{align*}
And similarly we have with probability at least $1-\delta - 5n^{-1} - 2\exp(-\Delta/\sigma)$
\begin{align*}
    L(TF_{\wha\theta}^+(\bfa H),\bfa P_1(\bfa z))\lesssim \exp\lef(-(1+o(1))\frac{\Delta^2}{8\sigma^2}\rig)+d\sqrt kn^{-1/4}C^d\sqrt{Polylog(n)+\log(2/\delta)}+N^{-100.5} .
\end{align*}

\section{Additional Experiments}

\begin{figure}[htbp]
    \centering
    \minipage{0.45\textwidth}
        \includegraphics[width=\linewidth]{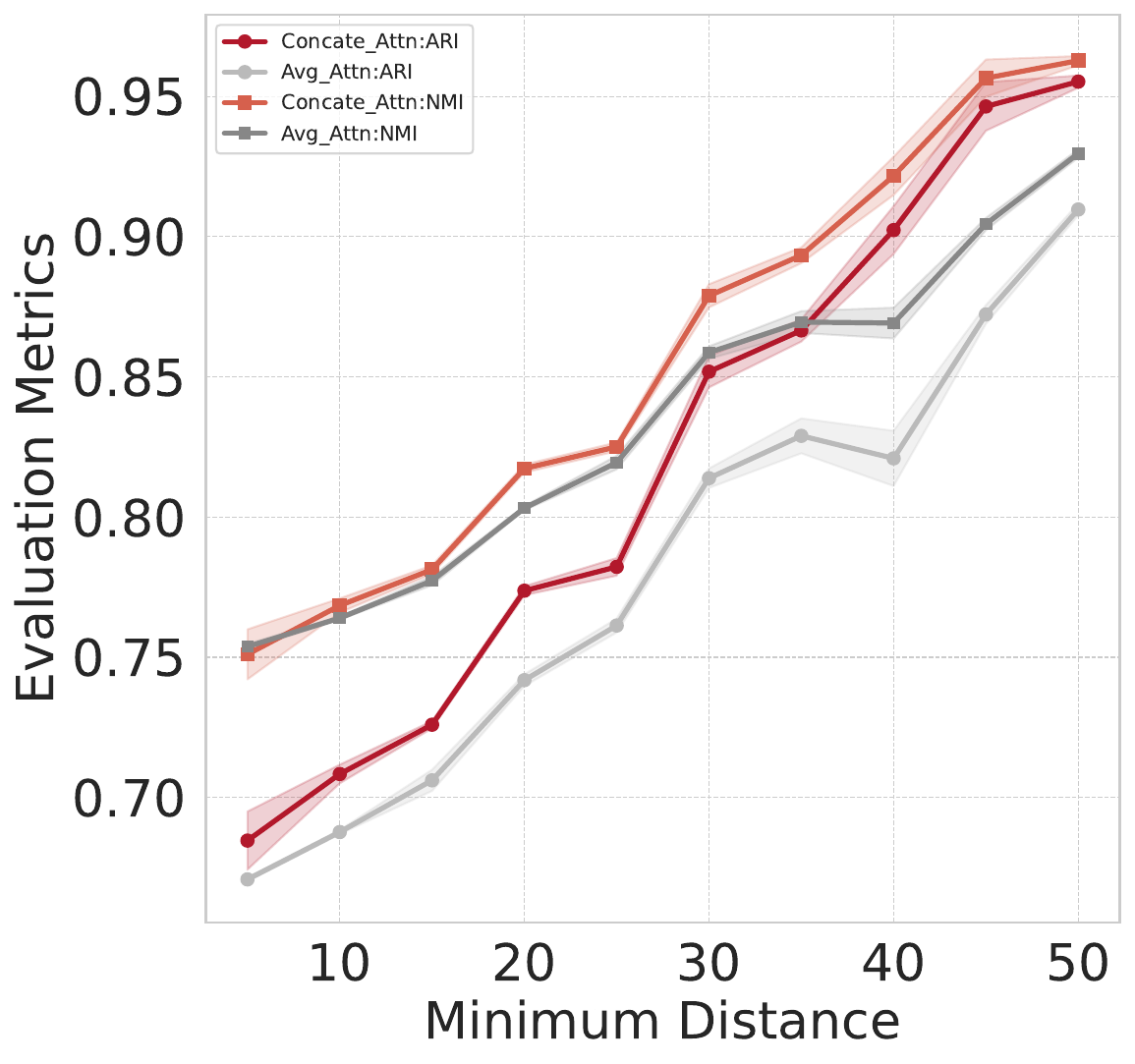}
        % \subcaption{Top-10 Eigenvalues}
    \endminipage
    \minipage{0.45\textwidth}
        \includegraphics[width=\linewidth]{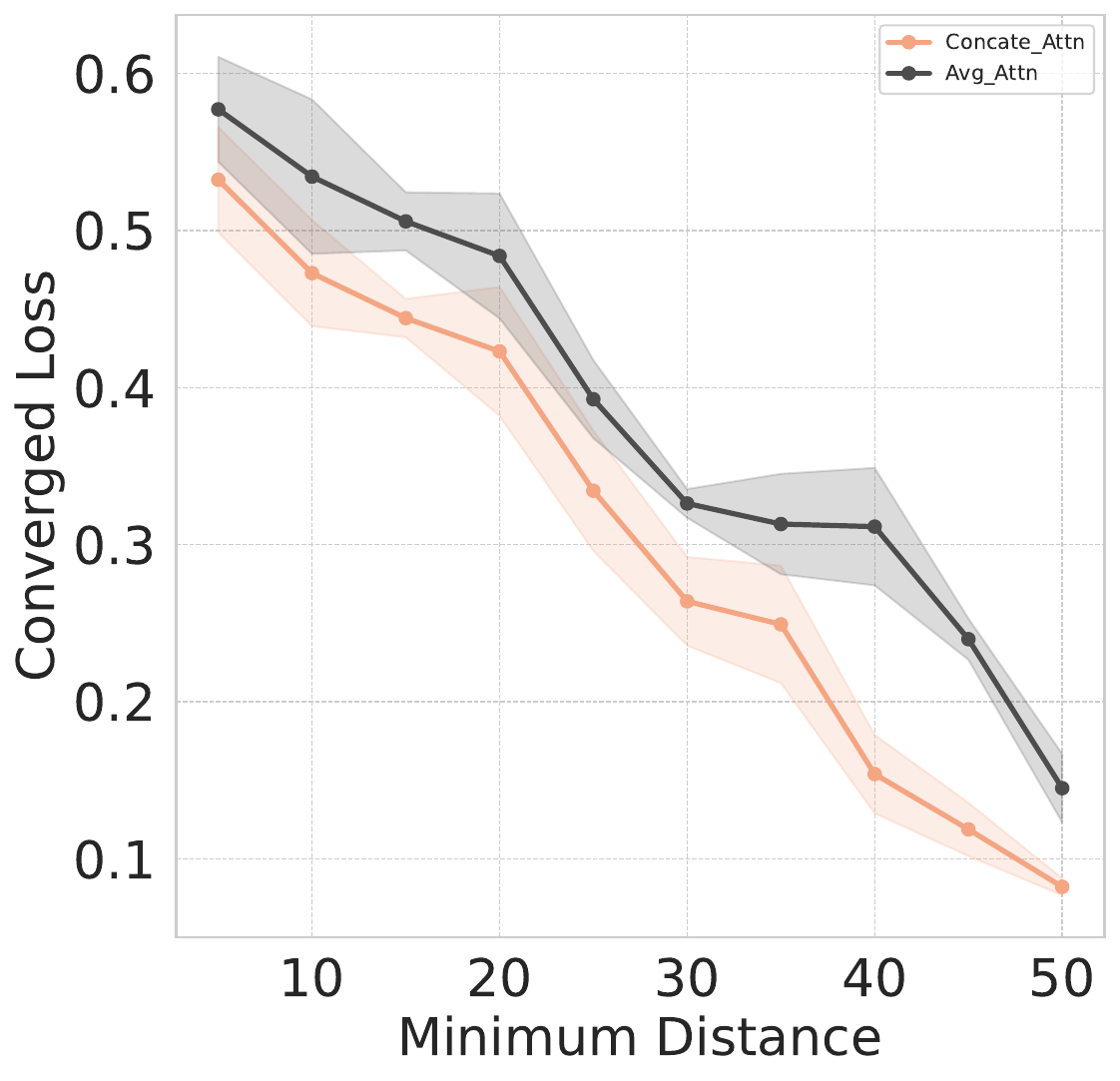}
        % \subcaption{RMSE with Varying Layers}
    \endminipage\vspace{-0.1em}
    \minipage{0.45\textwidth}
        \includegraphics[width=\linewidth]{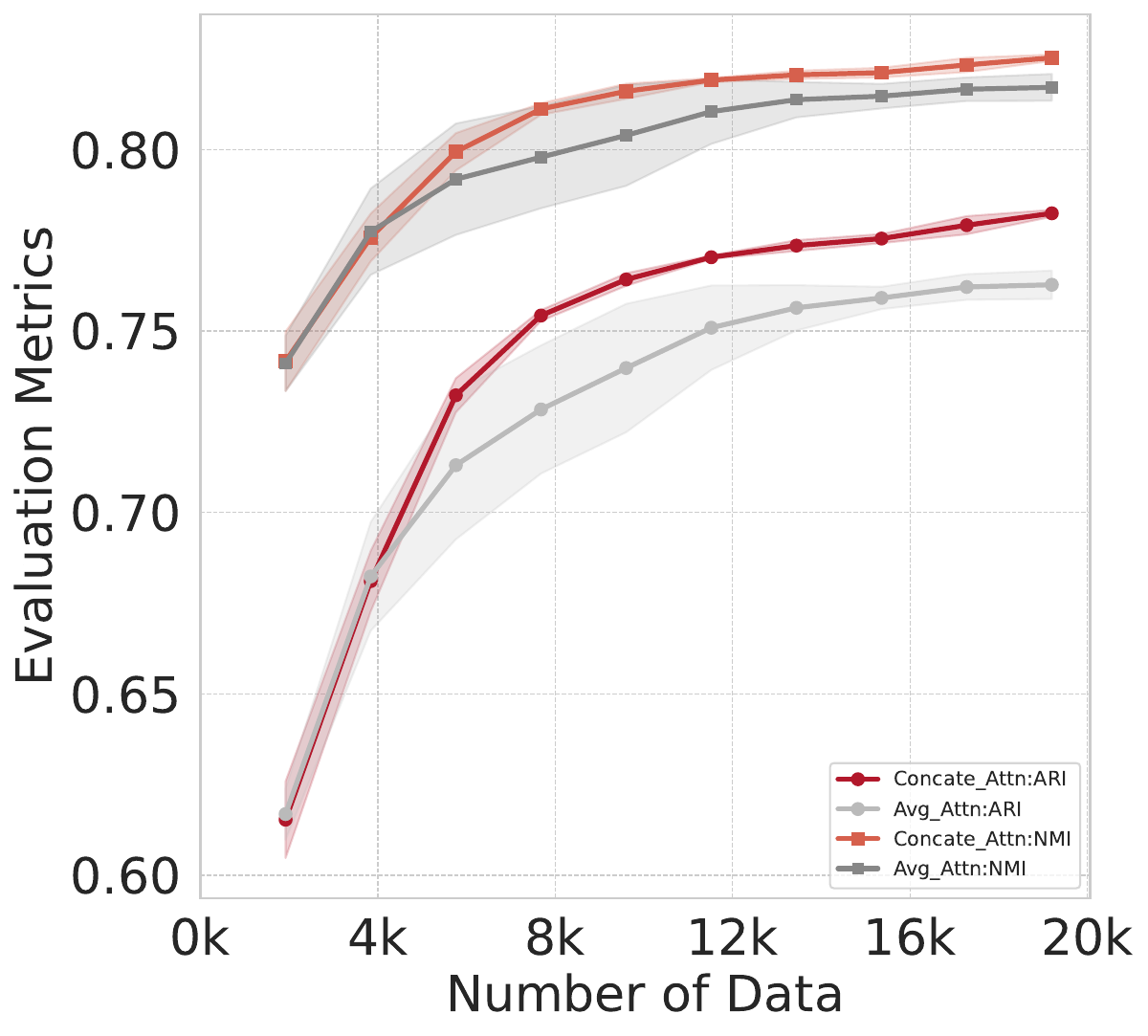}
        % \subcaption{Cosine Similarity with Varying d}
    \endminipage
    \minipage{0.45\textwidth}
        \includegraphics[width=\linewidth]{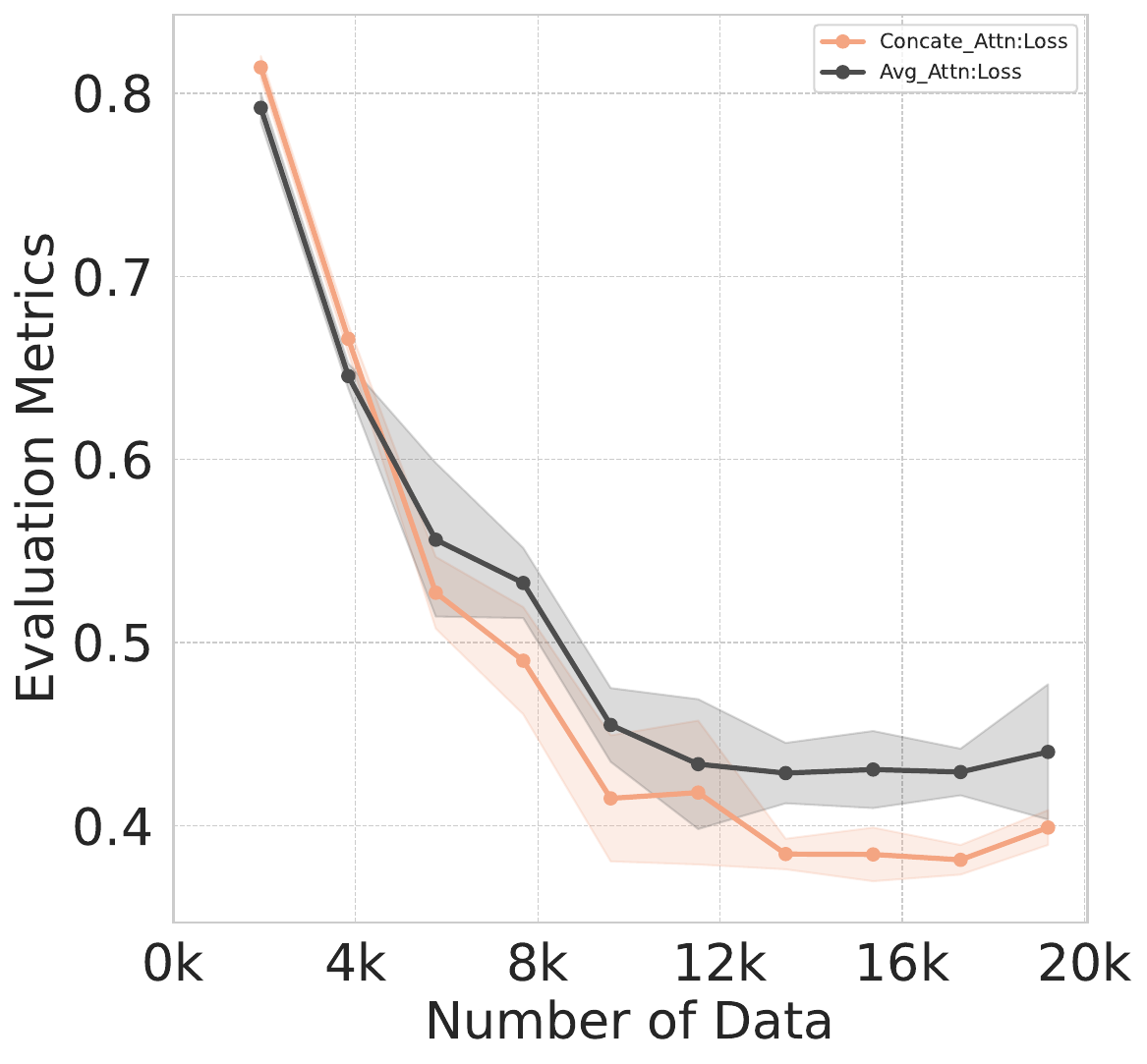}
        % \subcaption{Fourth Chart}
    \endminipage\vspace{-0.1em}
    \vspace{-1em}
    \caption{\textbf{Comparison of Concatenated Attention and Averaged Attention on Synthetic Dataset.}
    \emph{Top: Performance Comparision on Minimum Distance Task.}
    \emph{Bottom: Performance Comparision on Number of Data Task.}
    We observe similar trend of performance between concatenated multihead attention a averaged multihead attention across three tasks, two evaluation metrics, and the converged loss.
    All the experiment settings are the same as the experiments in \cref{sect4}.}
    \label{fig:attn_1}
\end{figure}

\begin{figure}[htbp]
    \centering
\minipage{0.45\textwidth}
        \includegraphics[width=\linewidth]{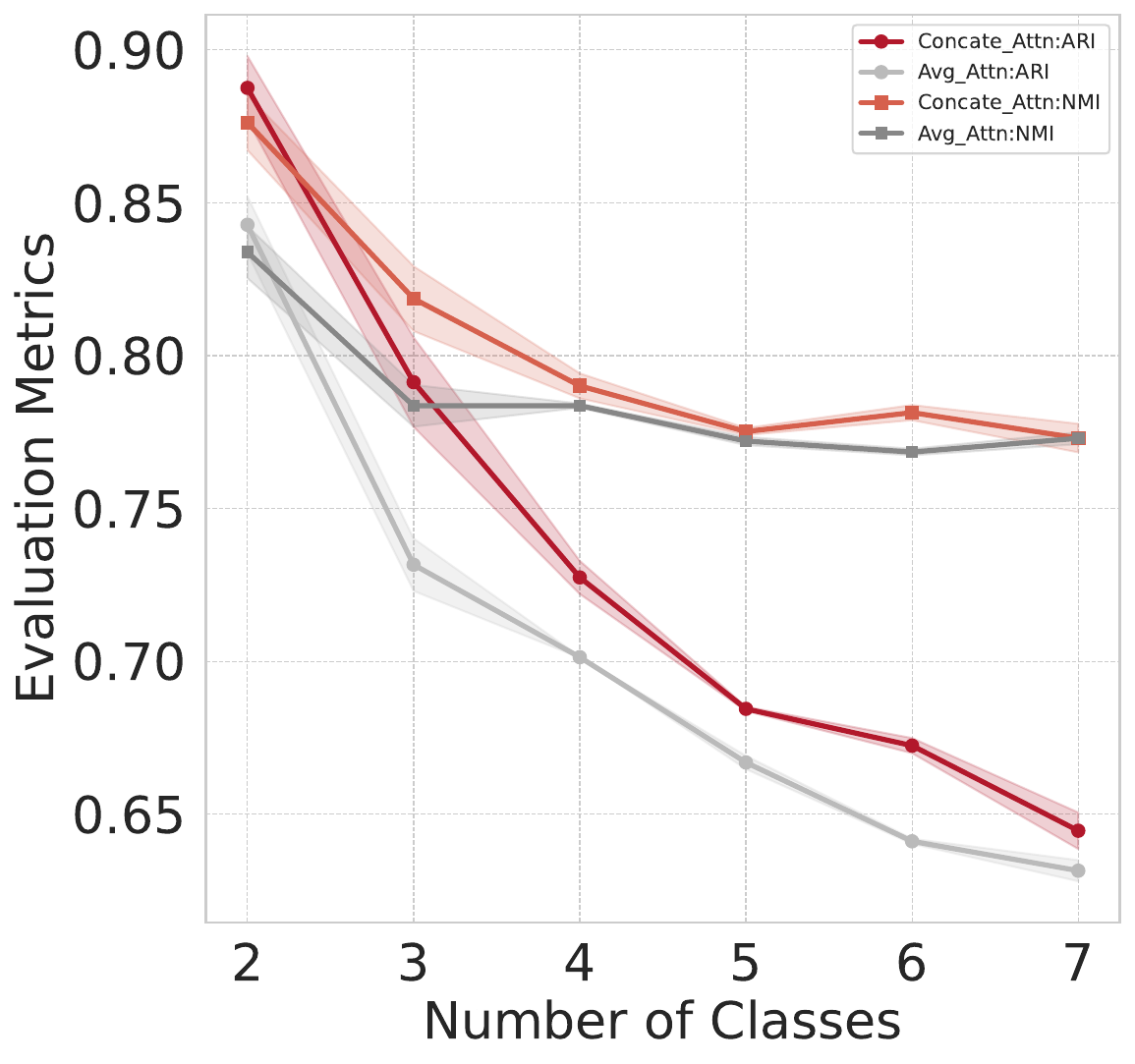}
        % \subcaption{Top-10 Eigenvalues}
    \endminipage
    \minipage{0.45\textwidth}
        \includegraphics[width=\linewidth]{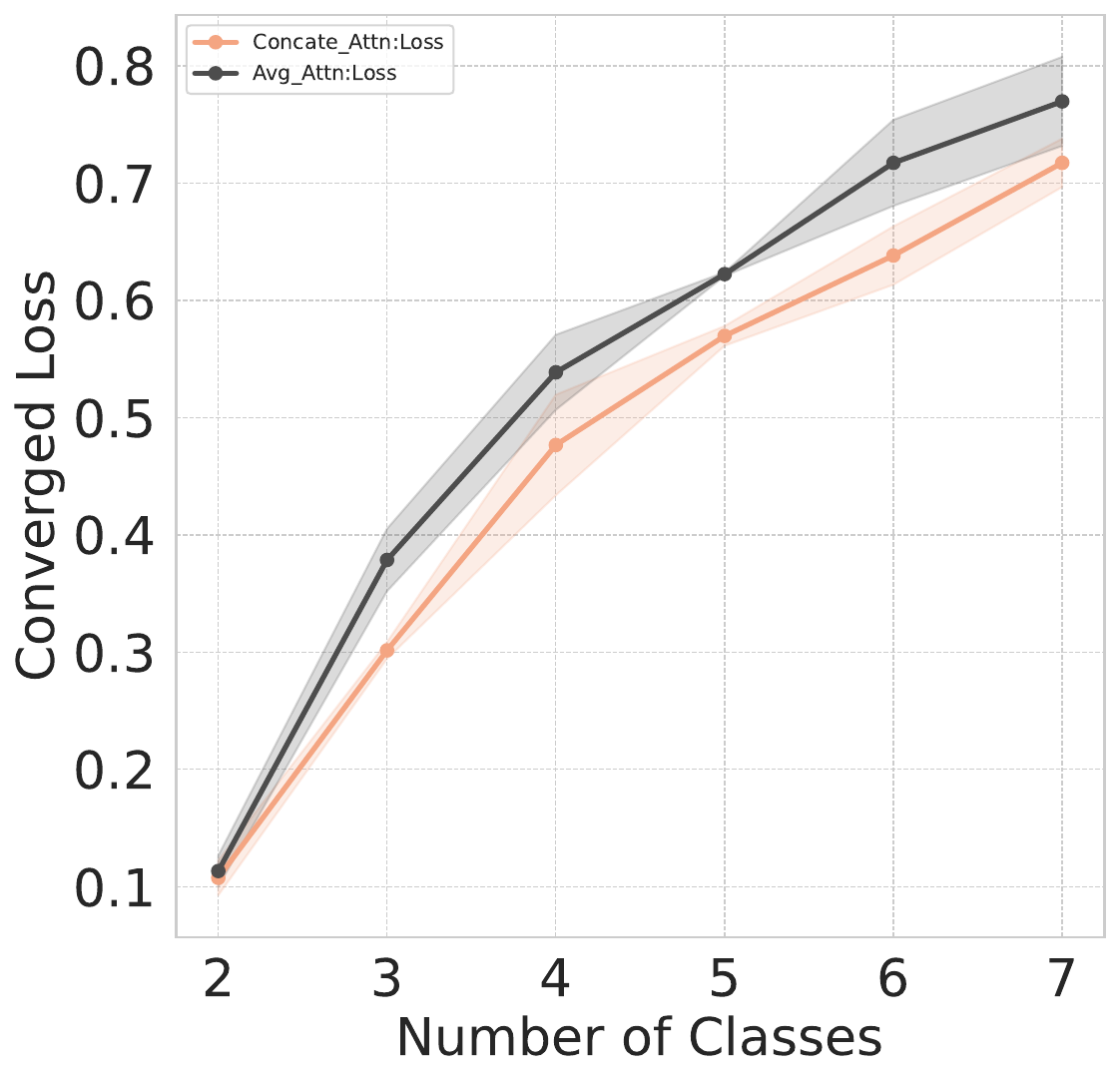}
        % \subcaption{RMSE with Varying Layers}
    \endminipage\vspace{-0.1em}
    \minipage{0.45\textwidth}
        \includegraphics[width=\linewidth]{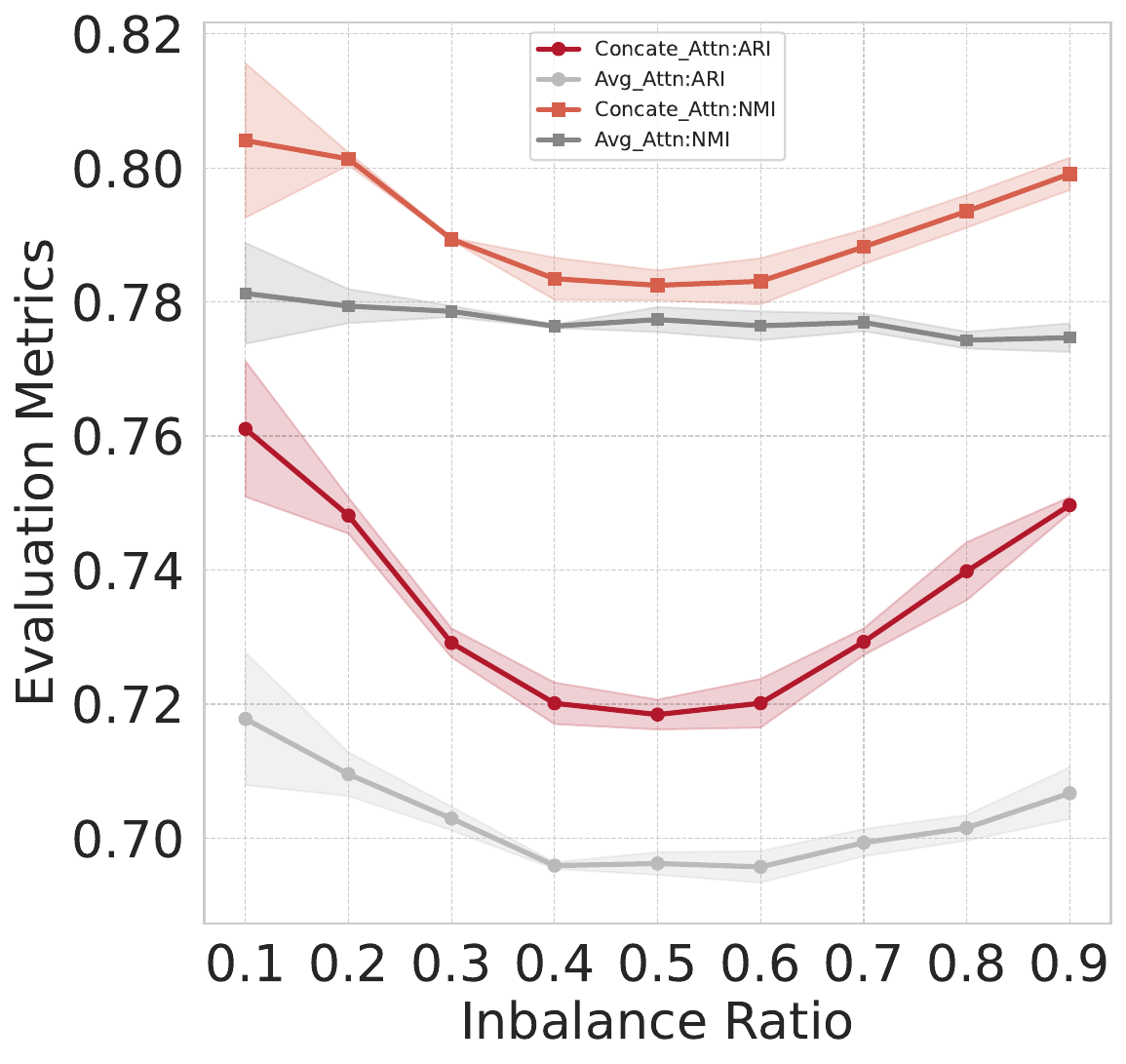}
        % \subcaption{Cosine Similarity with Varying d}
    \endminipage
    \minipage{0.45\textwidth}
        \includegraphics[width=\linewidth]{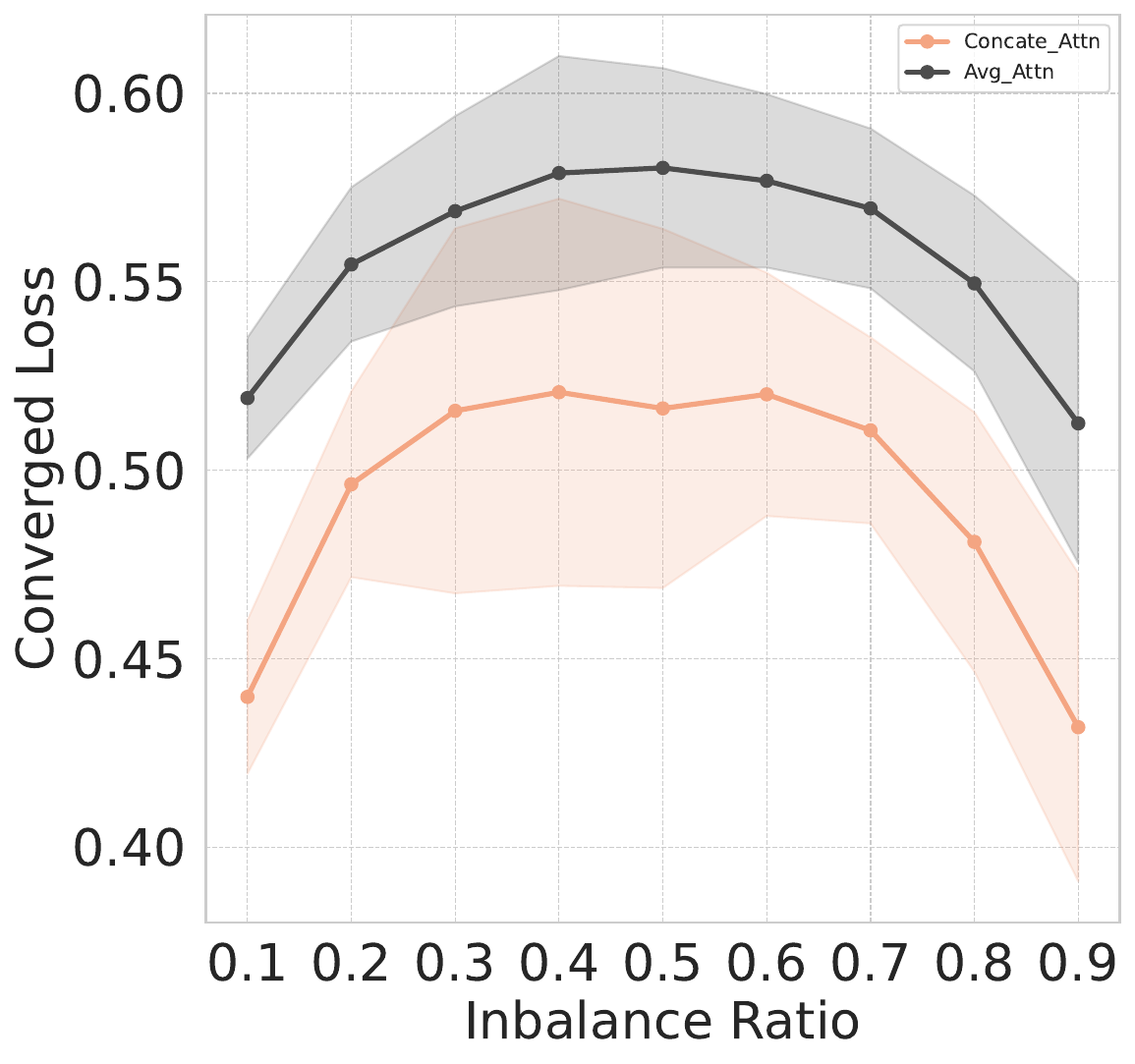}
        % \subcaption{Fourth Chart}
    \endminipage
    \vspace{-1em}
    \caption{{\textbf{Comparison of Concatenated Attention and Averaged Attention on Synthetic Dataset.}
    \emph{Top: Performance Comparision on Number of Classes Task.}
    \emph{Bottom: Performance Comparision on Inbalance Ratio Task.}
    }
    Again, we observe a similar performance trend between concatenated multihead attention and averaged multihead attention across both tasks.
    All experimental settings remain the same as those in \cref{sect4}.
    }
    % \label{fig:softmax_loss}
\end{figure}

\end{document}